\documentclass{ieeeaccess}
\usepackage{cite}
\usepackage{amsmath,amssymb,amsfonts}
\usepackage[T1]{fontenc}
\usepackage[utf8]{inputenc}
\usepackage{algorithmic}
\usepackage{graphicx}
\usepackage{textcomp}
\usepackage{float}
\usepackage{url}
\usepackage{bm}
\usepackage{booktabs} 
\usepackage{multirow} 
 \usepackage{hyperref}
\makeatletter
\AtBeginDocument{\DeclareMathVersion{bold}
\SetSymbolFont{operators}{bold}{T1}{times}{b}{n}
\SetSymbolFont{NewLetters}{bold}{T1}{times}{b}{it}
\SetMathAlphabet{\mathrm}{bold}{T1}{times}{b}{n}
\SetMathAlphabet{\mathit}{bold}{T1}{times}{b}{it}
\SetMathAlphabet{\mathbf}{bold}{T1}{times}{b}{n}
\SetMathAlphabet{\mathtt}{bold}{OT1}{pcr}{b}{n}
\SetSymbolFont{symbols}{bold}{OMS}{cmsy}{b}{n}
\renewcommand\boldmath{\@nomath\boldmath\mathversion{bold}}}
\makeatother

\def\BibTeX{{\rm B\kern-.05em{\sc i\kern-.025em b}\kern-.08em
    T\kern-.1667em\lower.7ex\hbox{E}\kern-.125emX}}

\begin{document}
\history{This work has been submitted to the IEEE for possible publication. Copyright may be transferred without notice, after which this version may no longer be accessible}
\doi{}

\title{Automated identification of Ichneumonoidea wasps via YOLO-based deep learning: Integrating HiresCam for Explainable AI}
\author{
    JOÃO MANOEL HERRERA PINHEIRO\authorrefmark{1} \IEEEmembership{Student Member, IEEE}, 
    GABRIELA DO NASCIMENTO HERRERA\authorrefmark{2}, 
    ALVARO DORIA DOS SANTOS\authorrefmark{3},
    LUCIANA BUENO DOS REIS FERNANDES\authorrefmark{2},
    RICARDO V. GODOY\authorrefmark{1},
    EDUARDO A. B. ALMEIDA\authorrefmark{4}, 
    HELENA CAROLINA ONODY\authorrefmark{5},
    MARCELO ANDRADE DA COSTA VIEIRA\authorrefmark{1},
    ANGÉLICA MARIA PENTEADO-DIAS\authorrefmark{2} AND
    MARCELO BECKER\authorrefmark{1} \IEEEmembership{Member, IEEE}
}

\address[1]{São Carlos School of Engineering, University of São Paulo, São Carlos 13566590, SP, Brazil}
\address[2]{Department of Ecology and Evolutionary Biology, Federal University of São Carlos, São Carlos 13565905, Brazil}
\address[3]{Federal University of Tocantins, Porto Nacional, 77500000, Brazil}
\address[4]{Department of Biology, University of São Paulo, Ribeirão Preto 14040901, Brazil}
\address[5]{State University of Piauí, Deputado Jesualdo Cavalcanti Campus, Corrente, 49800000, Brazil}

\tfootnote{This work was supported by the Petróleo Brasileiro S.A. - Petrobras, using resources from the R\&D clause of the ANP, in partnership with the Universidade de São Paulo (USP) and the Fundação de Apoio à Física e à Química (FAFQ), under Cooperation Agreement No. 2023/00016-6 and 2023/00013-7, Coordenação de Aperfeiçoamento de Pessoal de Nível Superior (CAPES) grant nº88887.002221/2024-00, nº88887.975224/2024–00 and (ROR identifier: 00x0ma614), Fundação de Amparo à Pesquisa do Estado de São Paulo (FAPESP) grant nº2014/50940-2, 2019/09215-6 and 2022/11451-2, Conselho Nacional de Desenvolvimento Científico e Tecnológico (CNPq) grant nº465562/2014-0, Instituto Nacional de Ciência e Tecnologia dos Hymenoptera Parasitoides (INCT-HYMPAR). For the purpose of open access, the authors have applied a Creative Commons CC BY license to any Author Accepted Manuscript version arising from this submission.}

\markboth
{J.M.H Pinheiro \headeretal: Automated identification of Ichneumonoidea wasps via YOLO-based deep learning}
{J.M.H Pinheiro \headeretal: Automated identification of Ichneumonoidea wasps via YOLO-based deep learning}

\corresp{CORRESPONDING AUTHOR: João Manoel Herrera Pinheiro (e-mail: joao.manoel.pinheiro@usp.br), Gabriela do Nascimento Herrera (e-mail: gabriela.herrera@estudante.ufscar.br
), Angélica Maria Pentado-Dias (e-mail: angelica@ufscar.br) and Marcelo Becker (e-mail: becker@sc.usp.br).}

\begin{abstract}
Accurate taxonomic identification of parasitoid wasps within the superfamily Ichneumonoidea is essential for biodiversity assessment, ecological monitoring, and biological control programs. However, morphological similarity, small body size, and fine-grained interspecific variation make manual identification labor-intensive and expertise-dependent. This study proposes a deep learning–based framework for the automated identification of Ichneumonoidea wasps using a YOLO-based architecture integrated with High-Resolution Class Activation Mapping (HiResCAM) to enhance interpretability. The proposed system simultaneously identifies wasp families from high-resolution images. The dataset comprises 3,556 high-resolution images of Hymenoptera specimens. The taxonomic distribution is primarily concentrated among the target families Ichneumonidae ($n=786$) and Braconidae ($n=648$), while other Hymenoptera families, such as Apidae ($n=466$) and Vespidae ($n=460$), were specifically included to increase generalization performance and improve the model's ability to discriminate parasitoids from related lineages. Extensive experiments were conducted using a curated dataset, with model performance evaluated through precision, recall, F1-score, and accuracy. The results demonstrate a high accuracy of over 96\% and robust generalization across morphological variations. HiResCAM visualizations confirm that the model focuses on taxonomically relevant anatomical regions, such as wing venation, antennae segmentation, and metasomal structures, thereby validating the biological plausibility of the learned features. The integration of explainable AI techniques improves transparency and trustworthiness. Consequently, this system functions as an effective automated triage tool, assisting taxonomists by performing the initial family-level separation of bulk samples and intelligently routing them to appropriate specialists, ultimately accelerating biodiversity characterization in this under-described parasitoid superfamily.
\end{abstract}


\titlepgskip=-21pt

\maketitle

\section{Introduction}
\label{sec:introduction}
\PARstart{I}{magine} inhabiting a world in which more than 80\% of species remain entirely unknown to science. This is the current state of our knowledge regarding Class Insecta~\cite{Camilo2011,2018Nigel}. Although insects represent the most species-rich group of animals and account for over half of all described species ~\cite{May1986,resh2009encyclopedia}, our inventory of this diversity is still far from complete. 

Approximately one million species have been formally described, and scientists estimate that an additional 5.5 million species remain undiscovered and undescribed ~\cite{2018Nigel,Eggleton2020}. We are currently facing a significant gap in insect taxonomy ~\cite{Eleanor2023,Ong2024}, a problem exacerbated by the ongoing global decline in insect species ~\cite{Wagner2021,Fenoglio2021}.

This decline has direct impacts on human well-being ~\cite{Schowalter2018}, as insects are a cornerstone of global biodiversity ~\cite{Cardoso2020} and perform crucial ecosystem functions. These functions include pollination ~\cite{GABRIEL200743}, maintaining the health of agricultural ecosystems ~\cite{Jankielsohn2018}, natural pest control ~\cite{Pardo2020}, and decomposition ~\cite{Eggleton2020}. 

Consequently, the accurate identification of insect species is vital for effective biodiversity monitoring and ecological research. Furthermore, precise classification is essential to distinguish agricultural pests from beneficial organisms. Contrary to common perception, the vast majority of insects are not harmful to humans ~\cite{Allison2023}

The order Hymenoptera comprises ants, bees, and wasps and represents one of the most species-rich insect orders ~\cite{Forbes2018}. Members of this order play essential ecological roles, particularly as pollinators ~\cite{Barbizan2009,Beggs2011}. Among Hymenoptera, the Ichneumonoidea superfamily is one of the most diverse in the Neotropics ~\cite{Quicke2015,Yu2016,yu2012world}. These wasps primarily parasitize larvae and pupae of holometabolous insects, although some groups can parasitize adult arthropods and arachnid oothecae, contributing to the maintenance of ecological balance ~\cite{Quicke2015}. The Ichneumonoidea superfamily comprises two major families, Ichneumonidae and Braconidae.

The Ichneumonidae, commonly known as Darwin wasps ~\cite{Klopfstein_2019}, is a hyper-diverse family of parasitoid wasps, with over 25,000 described species across 37 subfamilies and 1,450 genera ~\cite{yu2012world,Quicke2015}. Of these, 4,419 species have been described in the Neotropical region, and 955 have been recorded in Brazil. They are parasitoids of larvae and pupae of holometabolous insects, such as Coleoptera, Lepidoptera, and Hymenoptera, as well as other arthropods ~\cite{gauldHymenoptera1988,gauld_costa_rica}. Ichneumonidae have been comparatively less utilized in biological control programs, although their parasitoid behavior can effectively regulate the abundance of other insects, including agricultural pests ~\cite{Quicke2015}. Taxonomically, the group poses significant difficulties, as recognition of subfamilies is complex, particularly compared to that of the Braconidae. Identification is often restricted to females, as males frequently lack distinctive diagnostic features ~\cite{QuickeAsia2023}.

The Braconidae constitutes the second most diverse family within the Hymenoptera. This family includes over 21,000 described species across more than 1,100 genera, though these numbers represent only a fraction of their true global diversity ~\cite{yu2012world,Quicke2015,Chen2019}. Due to their prevalence as parasitoids of other insects ~\cite{Matthews1974}, braconids play a pivotal role in terrestrial ecosystems and are utilized as agents in biological control programs ~\cite{shaw1991classification}. Taxonomically, the most reliable distinction is found in the wing venation: braconids almost invariably lack the second recurrent vein (2m-cu) in the fore wing, a vein that is typically present in ichneumonids ~\cite{Quicke2015}.

Traditionally, insect identification has been the domain of taxonomist, relying heavily on morphological examination under microscopes, detailed dichotomous keys, and extensive reference collections~\cite{Benjamin2016}. This classical approach, while foundational to our understanding of insect diversity, is inherently labor-intensive, time-consuming, and demands highly specialized training and years of experience~\cite{Magni2023}.

In this study, we present a novel deep learning framework specifically designed for the automated identification of the hyper-diverse Ichneumonoidea superfamily, representing the first dedicated computational approach focused on this taxonomically complex group. By leveraging transfer learning and benchmarking state-of-the-art architectures, including YOLOv12 and YOLOv26, our proposed workflow functions as an operational pipeline for automated taxonomic triage.

\begin{figure}[ht]
    \centering
        \includegraphics[width=0.8\linewidth]{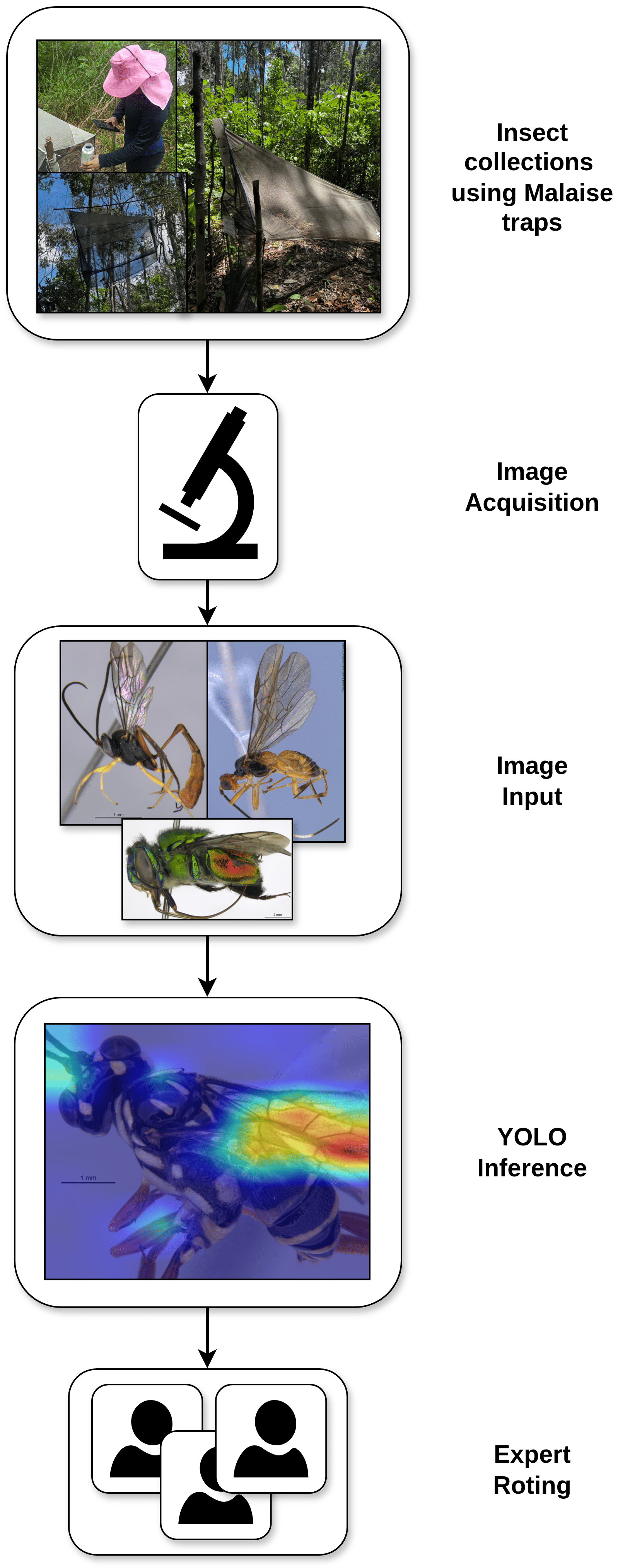}
	\caption{Proposed operational framework for the automated taxonomic triage of parasitoid wasps. The end-to-end workflow comprises five sequential stages: (1) field collection of insects using Malaise traps, (2) laboratory image acquisition, (3) image input, (4) YOLO-based inference coupled with HiResCAM visual interpretation, and (5) expert routing. By filtering bulk samples at the family level, the system intelligently directs Ichneumonidae and Braconidae specimens to their respective taxonomic specialists. This pre-sorting mechanism optimizes human expertise by eliminating the manual burden of preliminary sorting.}
	\label{fig_automated_framework}
\end{figure}

Crucially, this system assists taxonomists by performing the initial separation of bulk samples at the family level, intelligently routing specimens directly to the correct specialists (e.g., directing Braconidae and Ichneumonidae samples to their respective experts). This pre-sorting mechanism eliminates the manual burden of preliminary sorting and optimizes the allocation of specialized human expertise as detaild in Figure \ref{fig_automated_framework}.

A critical component of this approach is the integration of Explainable Artificial Intelligence (XAI) techniques, specifically HiResCAM, which provides high-resolution visual interpretations of the model's internal decision-making process. These visualizations enable the identification of morphologically relevant regions, such as wing venation and metasomal structures, that align with traditional taxonomic criteria, thereby enhancing the transparency and biological plausibility of the predictions. The dataset and source code are publicly available to ensure reproducibility and to support further research in biodiversity informatics.

The remainder of this paper is organized as follows: Section \ref{sec:related} reviews related work in the automated identification of parasitoid wasps, highlighting the current lack of dedicated computational studies focusing on the Ichneumonoidea superfamily. Section \ref{sec:methods} details the proposed methodology, including the dataset characteristics and the deep learning architectures utilized. Section \ref{sec:results} presents the experimental results and provides a comprehensive discussion of the findings, comparing the model's visual focus with traditional morphological keys. Finally, Section \ref{sec:conclusion} concludes the paper by summarizing the main contributions, addressing the study's limitations, and outlining directions for future research.

\section{Related work}
\label{sec:related}
Deep learning, a rapidly evolving field within artificial intelligence, utilizes computational models composed of multiple processing layers to learn abstract data representations ~\cite{goodfellow2016deep,bishop2023deep}. This distinguishes deep learning from traditional statistical prediction approaches ~\cite{Sarker2021}. These methods have significantly advanced various domains, including image classification, semantic segmentation, object detection, and speech recognition ~\cite{Shinde2018,Sharifani2023}. 

The core principle involves discovering intricate structures in large datasets through the backpropagation algorithm, which dictates how a machine adjusts its internal parameters to compute representations across layers ~\cite{LeCun2015,Zhao2024}. Unlike traditional machine learning that relies on carefully engineered feature extractors, deep learning automatically discovers the necessary representations from raw data ~\cite{Mahony2020,Procedia2020} and has shown promising results across several application domains ~\cite{Alzubaidi2021,Bhatt2021}.

While deep learning has received significant attention in other domains, its application in invertebrate monitoring and biodiversity research has been slow to develop ~\cite{Christin2019}. However, this has changed over the past decade, as deep learning has begun to revolutionize the fields of entomology and ecology ~\cite{Weinstein2018,Wenyong2021,Toke2021}. 

Deep learning and computer vision offer potential solutions to the long-standing challenges of inefficient and labor intensive insect identification ~\cite{JohannaArje2020,Telmo2020,Teixeira2023,Yuanyi2024}, monitoring ~\cite{Simon2024}, and pest detection ~\cite{Wu2019,Barbedo2020,Batz2023,Passias2024}. However, a critical gap remains in the literature regarding the taxonomic complexity of the Ichneumonoidea superfamily.

In the context of evaluating automated identification systems for hyper-diverse taxa, it is notable that some studies, such as the DiversityScanner ~\cite{Wuhrl2022,Wuhrl2024}, have detected and identified 14 families for robot handling with a precision of 91.4\%. However, in their evaluation of the Ichneumonoidea superfamily, only 246 images were used to assess the identification model's performance. Furthermore, while the system employed class activation maps to visualize the features the neural network prioritized during identification, these heatmaps were primarily used for internal model validation rather than being systematically compared against established morphological keys.

In the domain of parasitoid wasps, ~\cite{Shirali2024} demonstrated the efficacy of deep learning for identifying the highly diverse and cryptic Diapriidae family, using a dataset of 2,257 images, their study compared three architectures, with the BEiTv2 transformer model achieving the highest accuracy of 96\% for genus-level identification and 97\% for sex determination, significantly outperforming YOLOv8 and ConvNeXt.

\section{Materials and methods}
\label{sec:methods}
\subsection{Data collection}
\begin{figure}[H]
    \centering
        \includegraphics[width=0.6\linewidth]{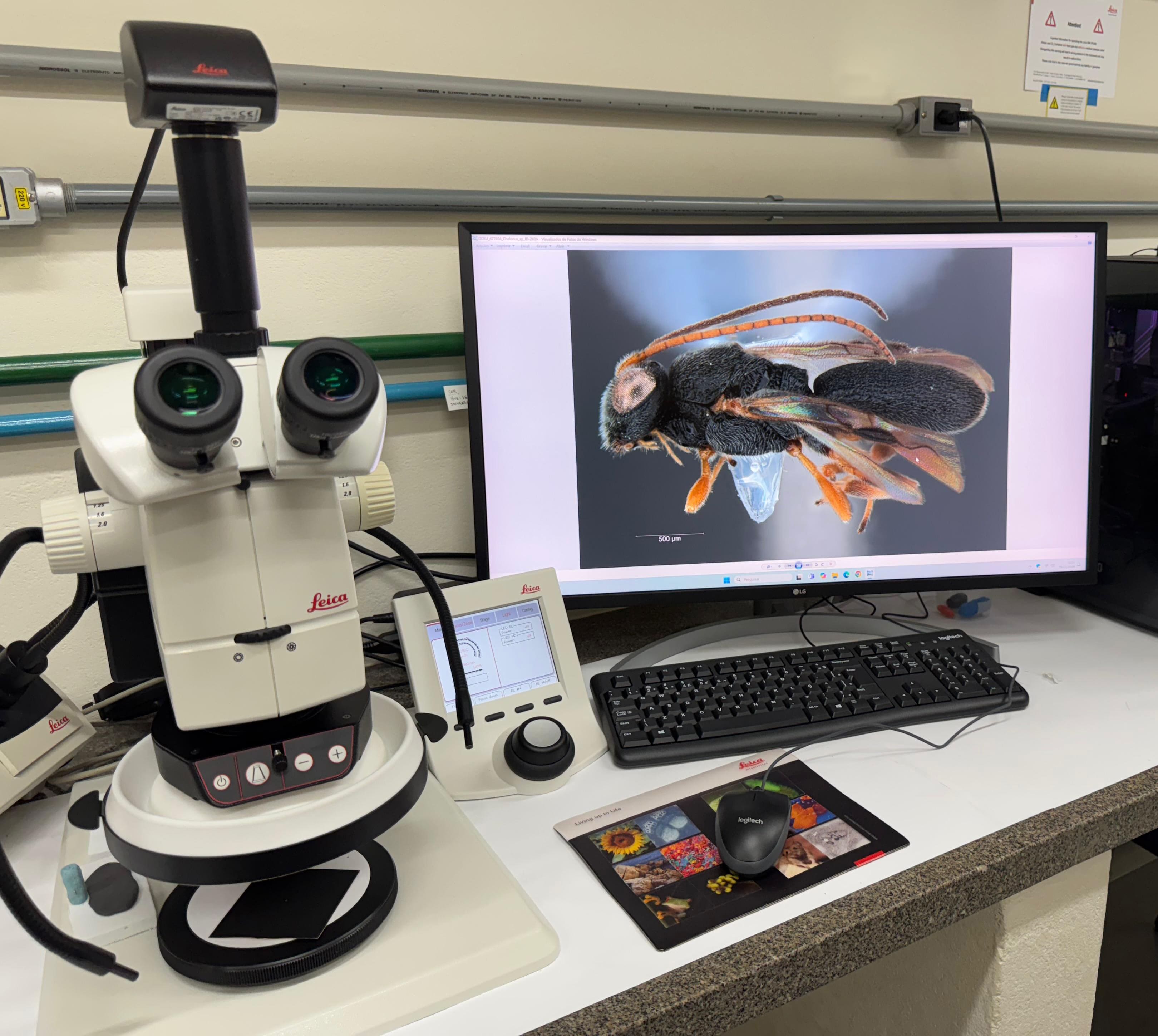}
	\caption{Specimens were retrieved from DCBU collection ~\cite{dias2025dcbu}. Imaged from ~\cite{Pinheiro11483097}.}
	\label{fig_species}
\end{figure}

\begin{figure*}{}
    \centering
    % --- Row 1 ---
    \begin{minipage}[t]{0.25\textwidth}
        \centering
        \includegraphics[width=\linewidth]{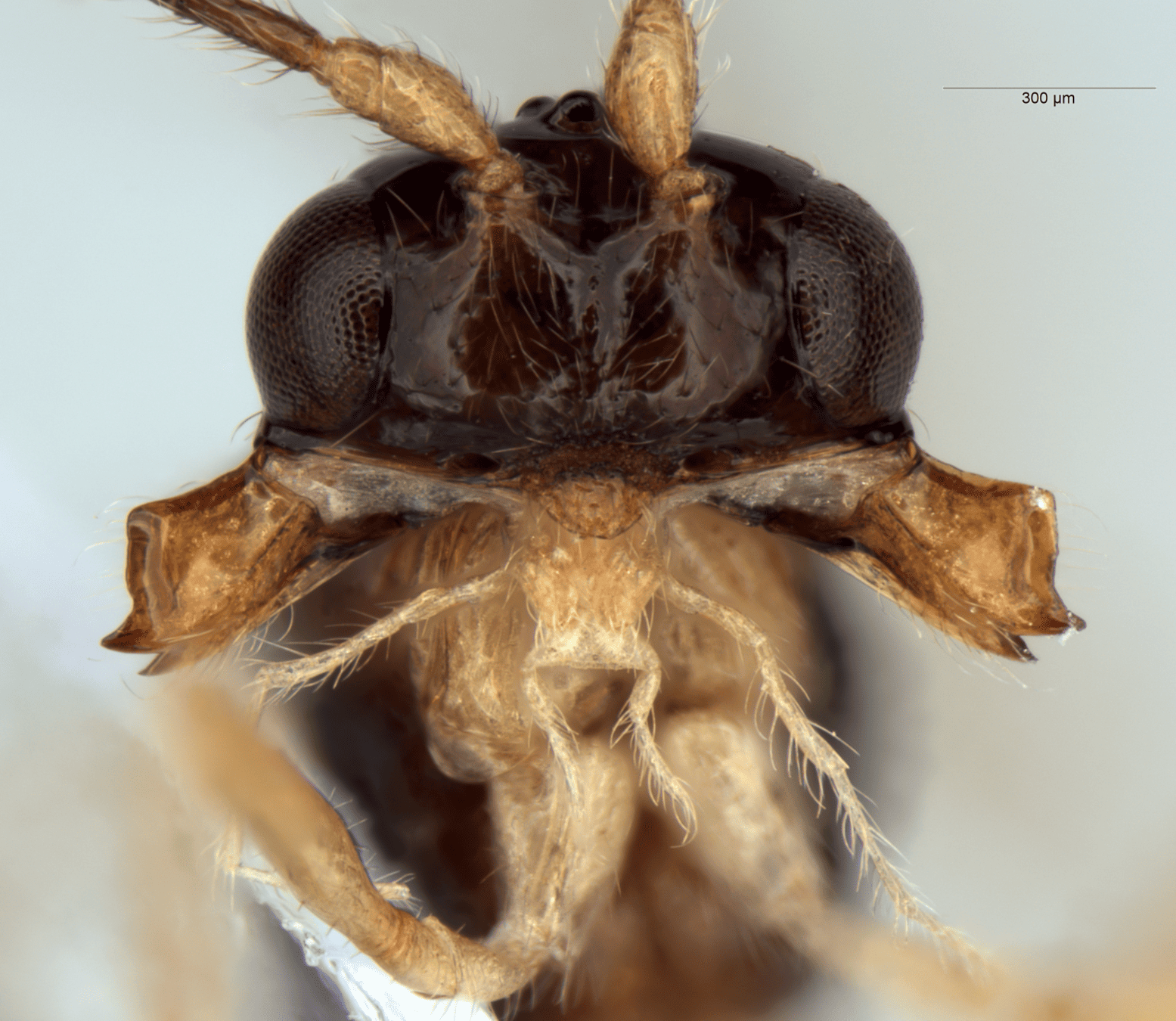}
        \par (a)
    \end{minipage}
    \hfill
    \begin{minipage}[t]{0.3\textwidth}
        \centering
        \includegraphics[width=\linewidth]{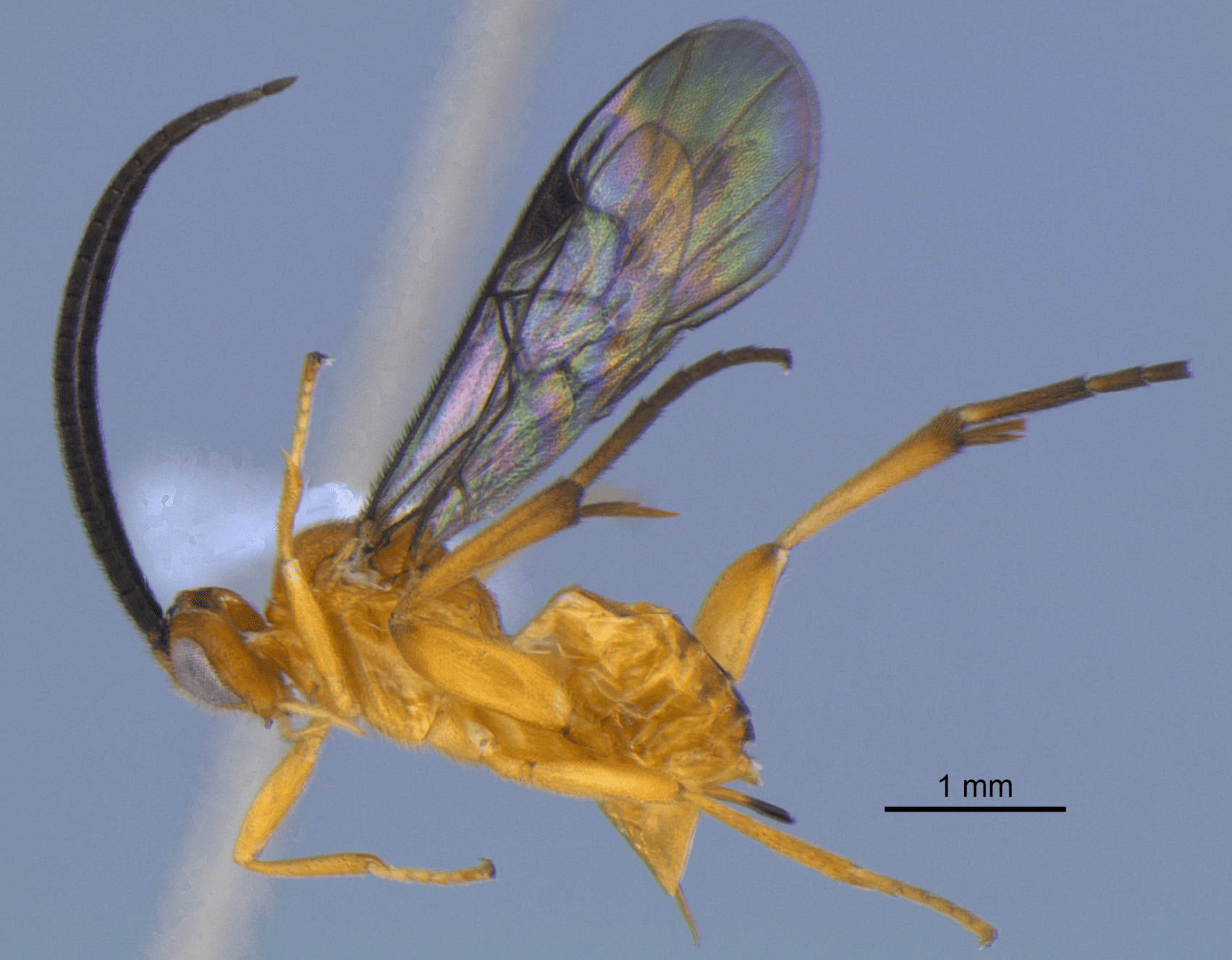}
        \par (b)
    \end{minipage}
    \hfill
        \begin{minipage}[t]{0.35\textwidth}
        \centering
        \includegraphics[width=\linewidth]{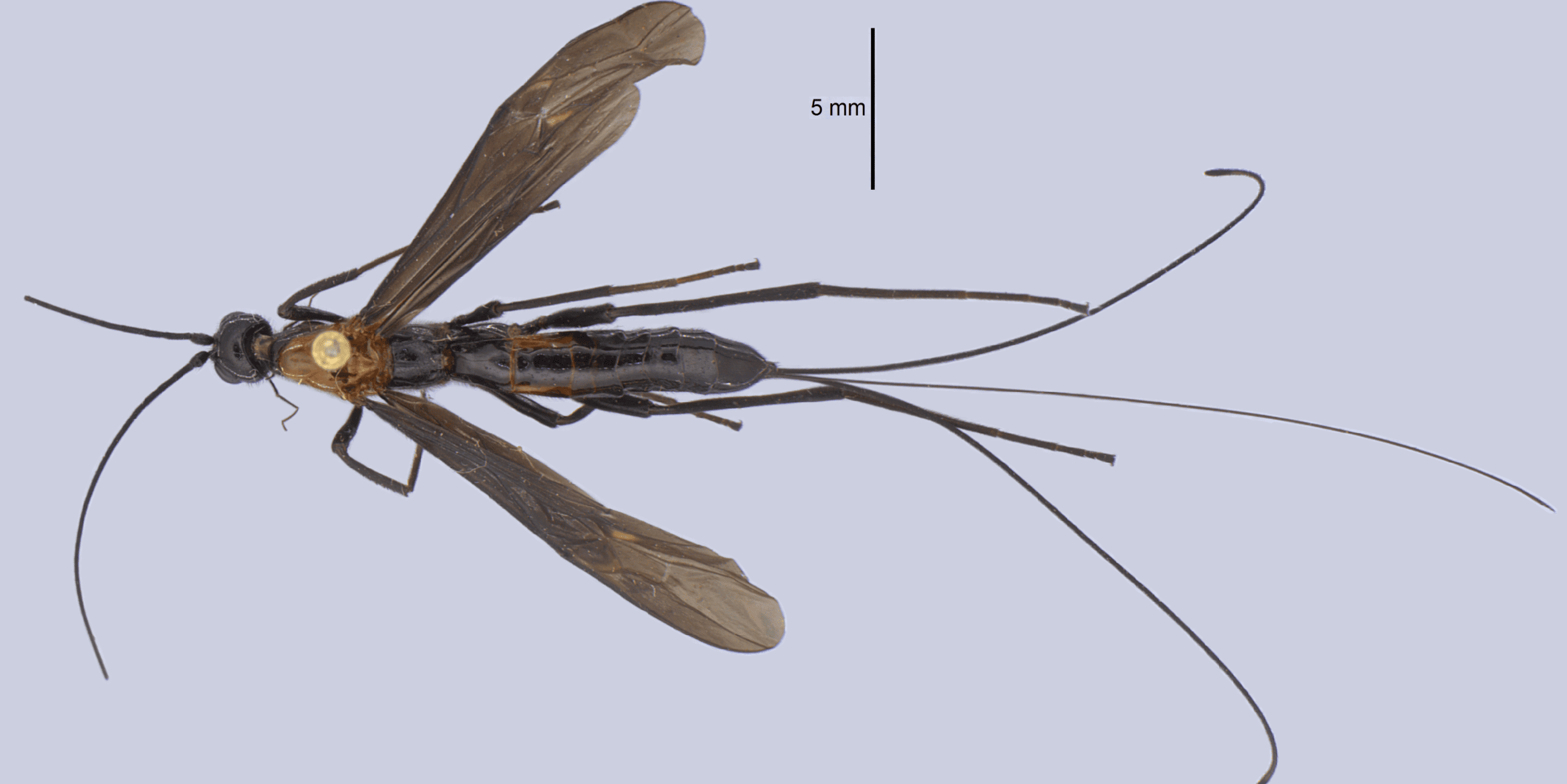}
        \par (c)
    \end{minipage}
    \vspace{0.1cm} % Add some vertical spacing between rows

    % --- Row 2 ---
    \begin{minipage}[t]{0.3\textwidth}
        \centering
        \includegraphics[width=\linewidth]{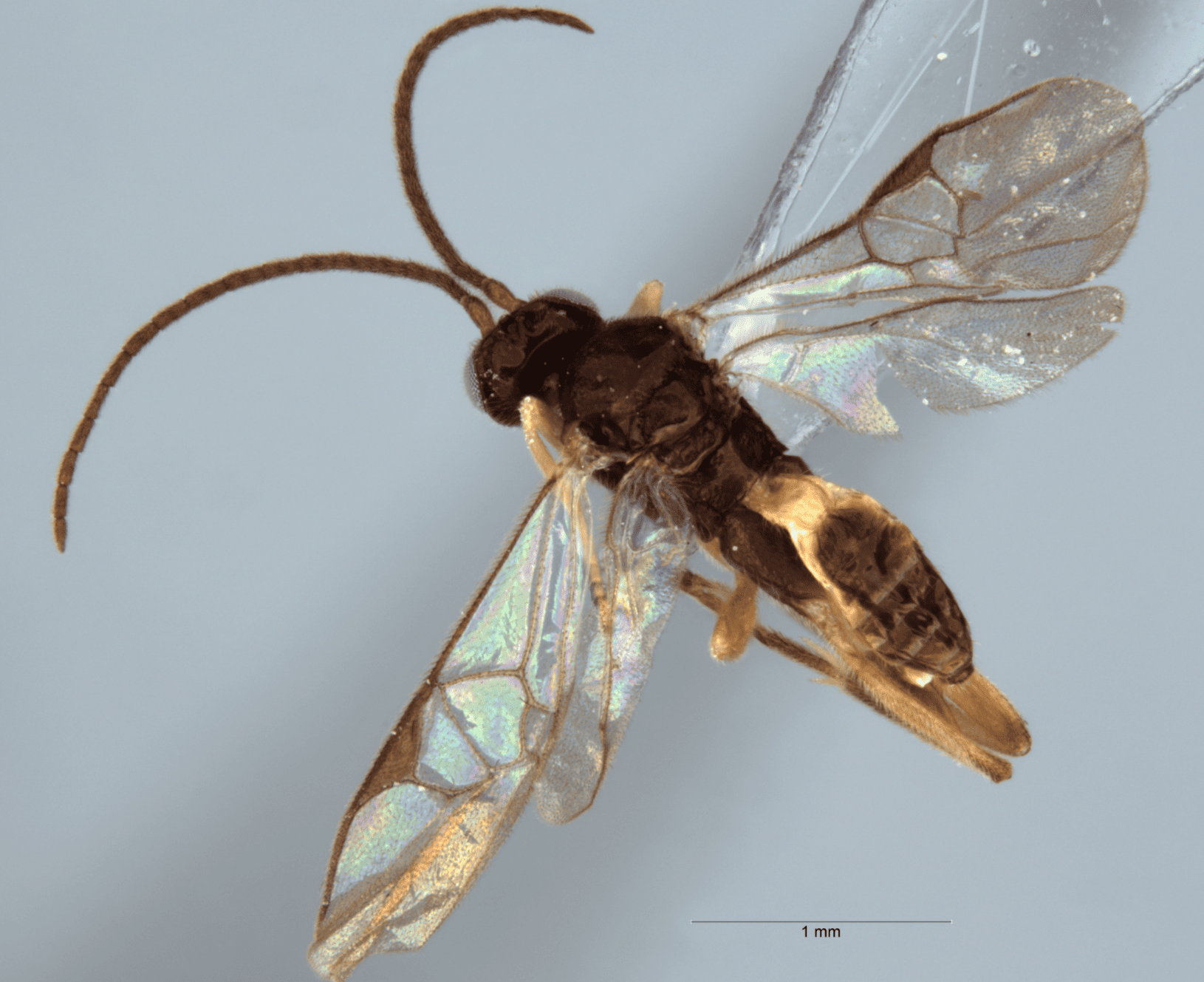}
        \par (d)
    \end{minipage}
    \hfill
        \begin{minipage}[t]{0.25\textwidth}
        \centering
        \includegraphics[width=\linewidth]{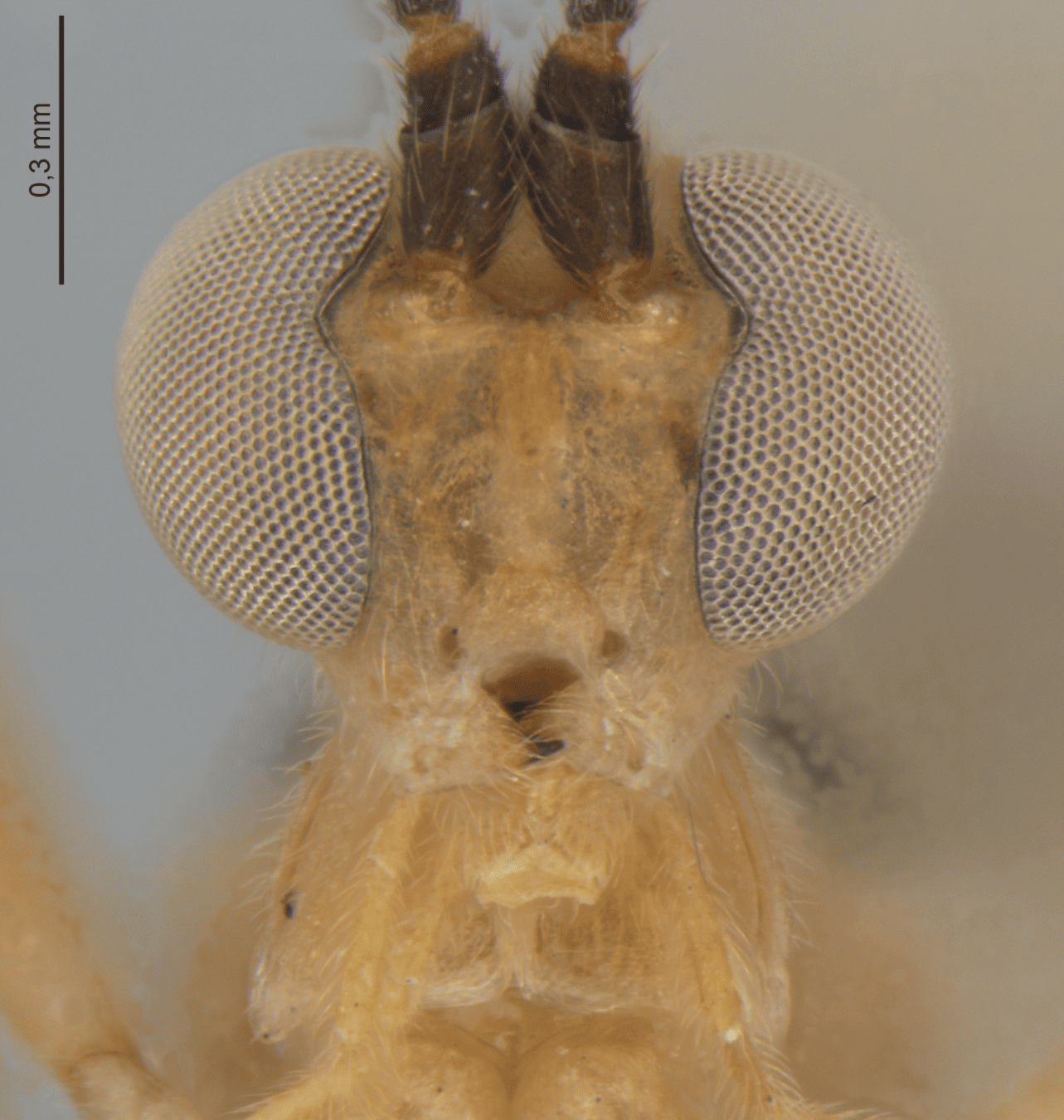}
        \par (e)
    \end{minipage}
    \hfill
        \hfill
        \begin{minipage}[t]{0.35\textwidth}
        \centering
        \includegraphics[width=\linewidth]{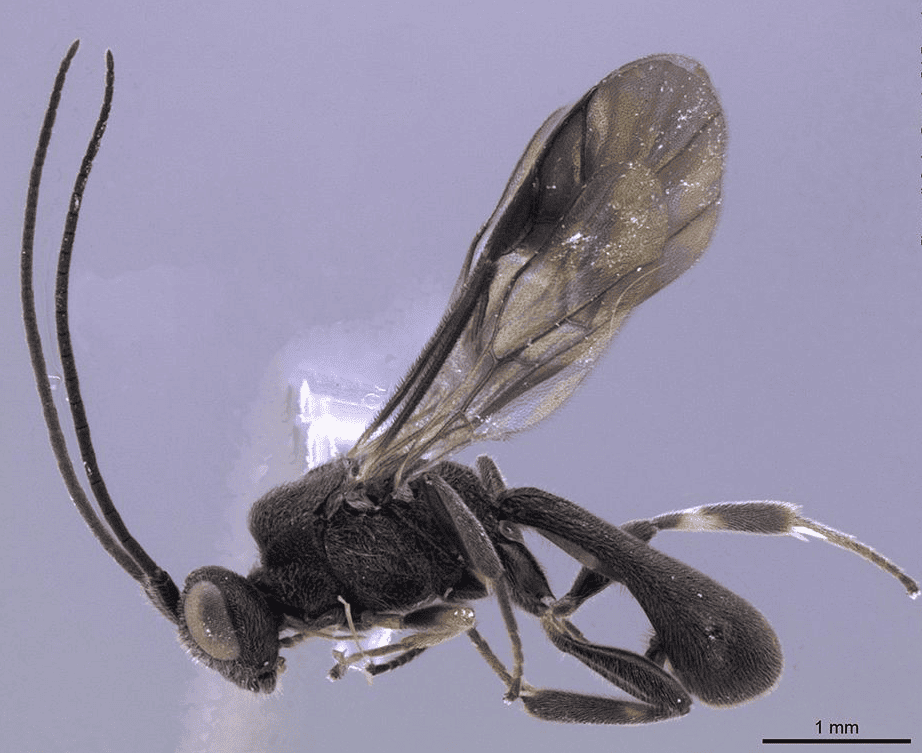}
        \par (f)
    \end{minipage}

    \vspace{0.1cm} % Add some vertical spacing between rows

    % --- Row 3 ---
    \begin{minipage}[t]{0.3\textwidth}
        \centering
        \includegraphics[width=\linewidth]{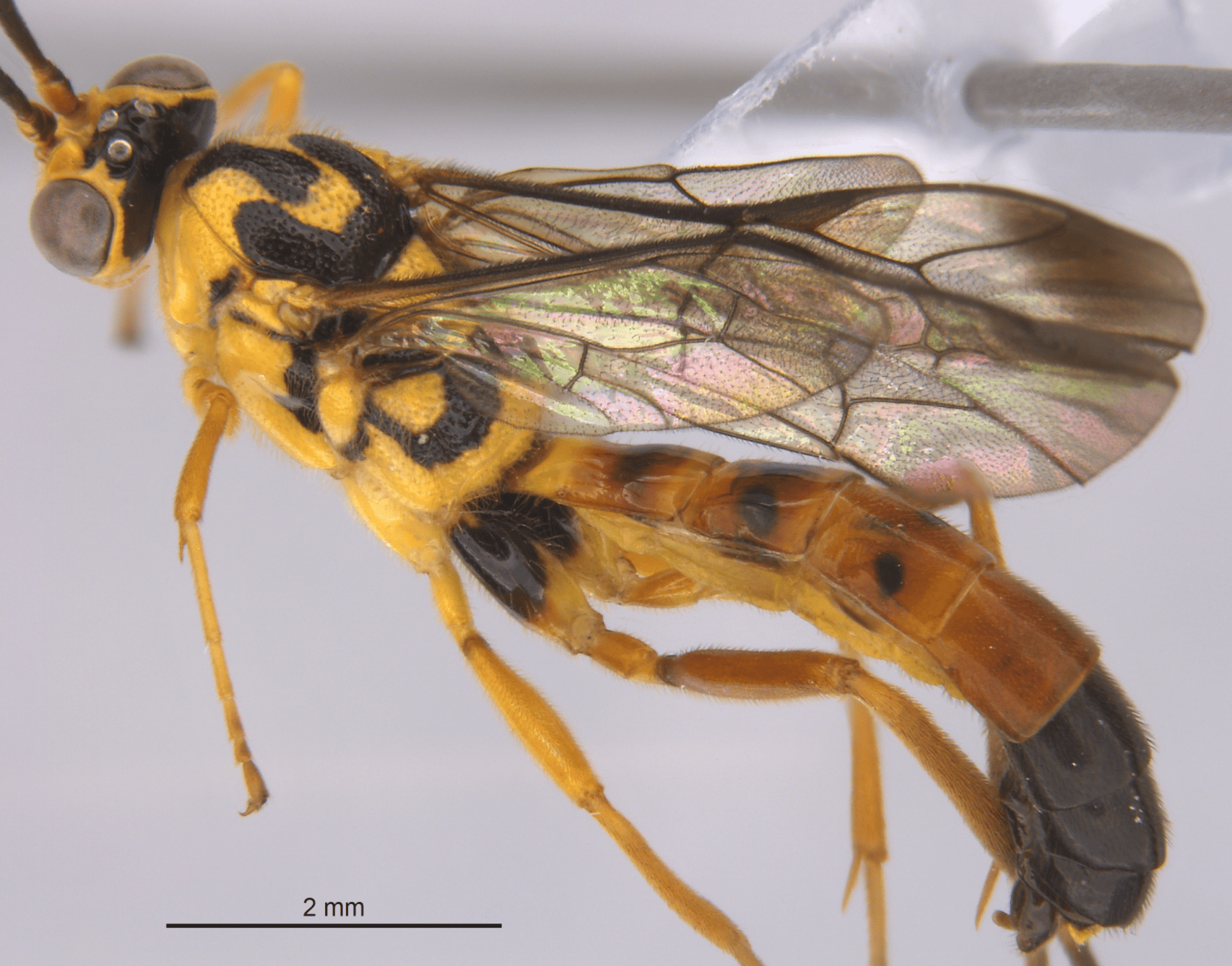}
        \par (g)
    \end{minipage}
    \hfill
    \begin{minipage}[t]{0.3\textwidth}
    \centering
    \includegraphics[width=\linewidth]{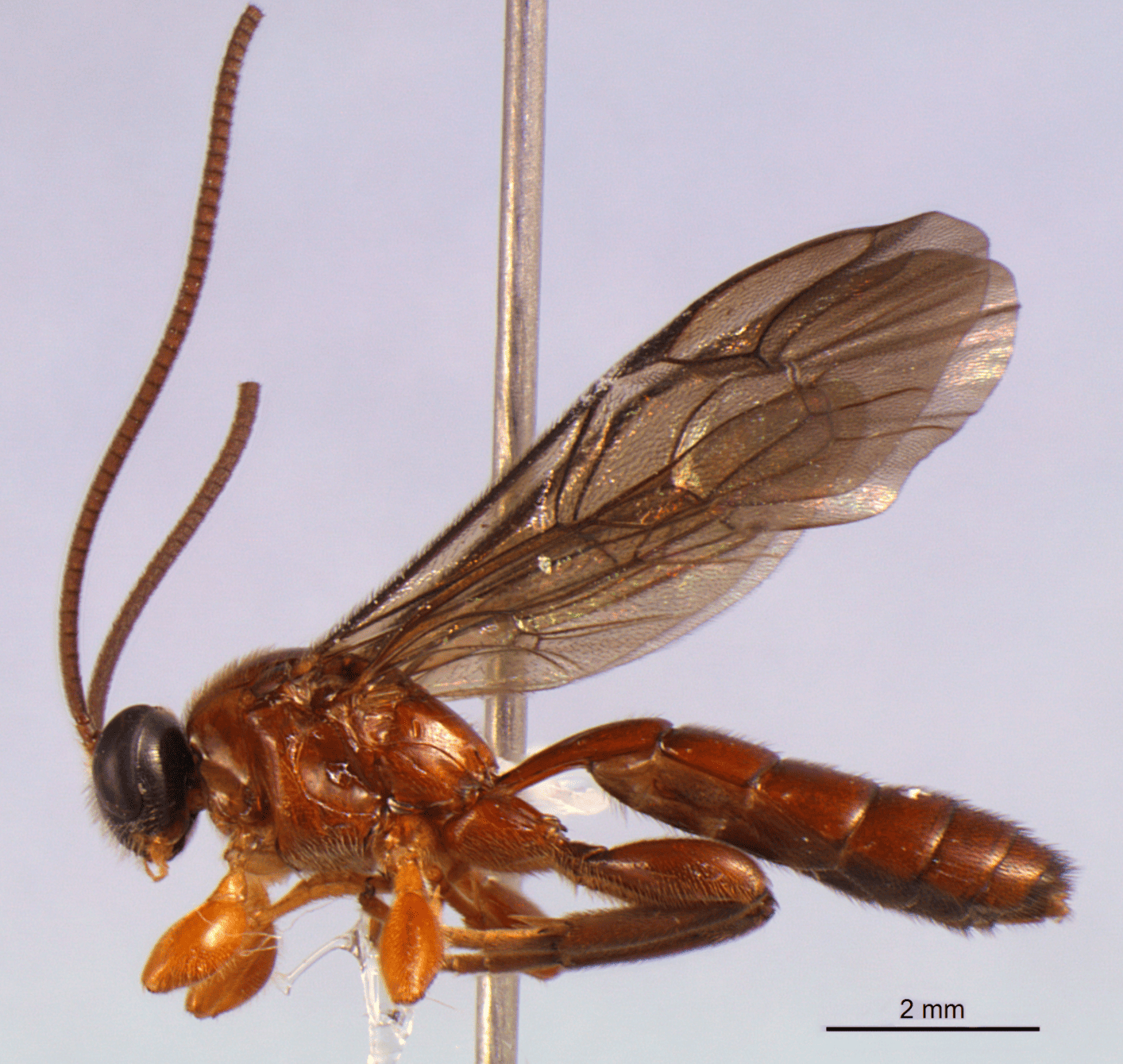}
    \par (h)
    \end{minipage}
    \hfill
    \begin{minipage}[t]{0.3\textwidth}
        \centering
        \includegraphics[width=\linewidth]{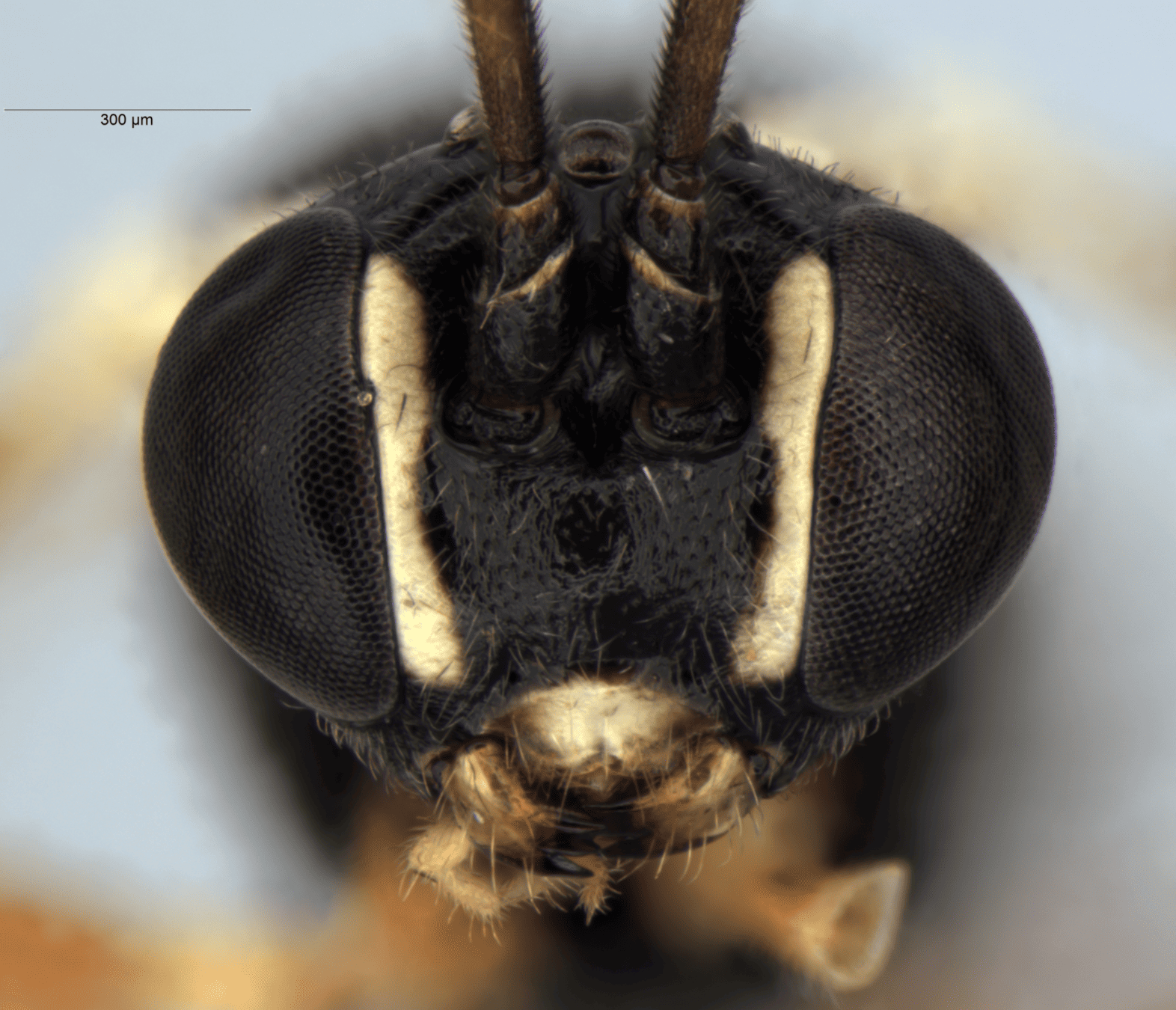}
        \par (i)
    \end{minipage}

    \vspace{0.1cm} % Add some vertical spacing between rows
    
    \begin{minipage}[t]{0.25\textwidth}
        \centering
        \includegraphics[width=\linewidth]{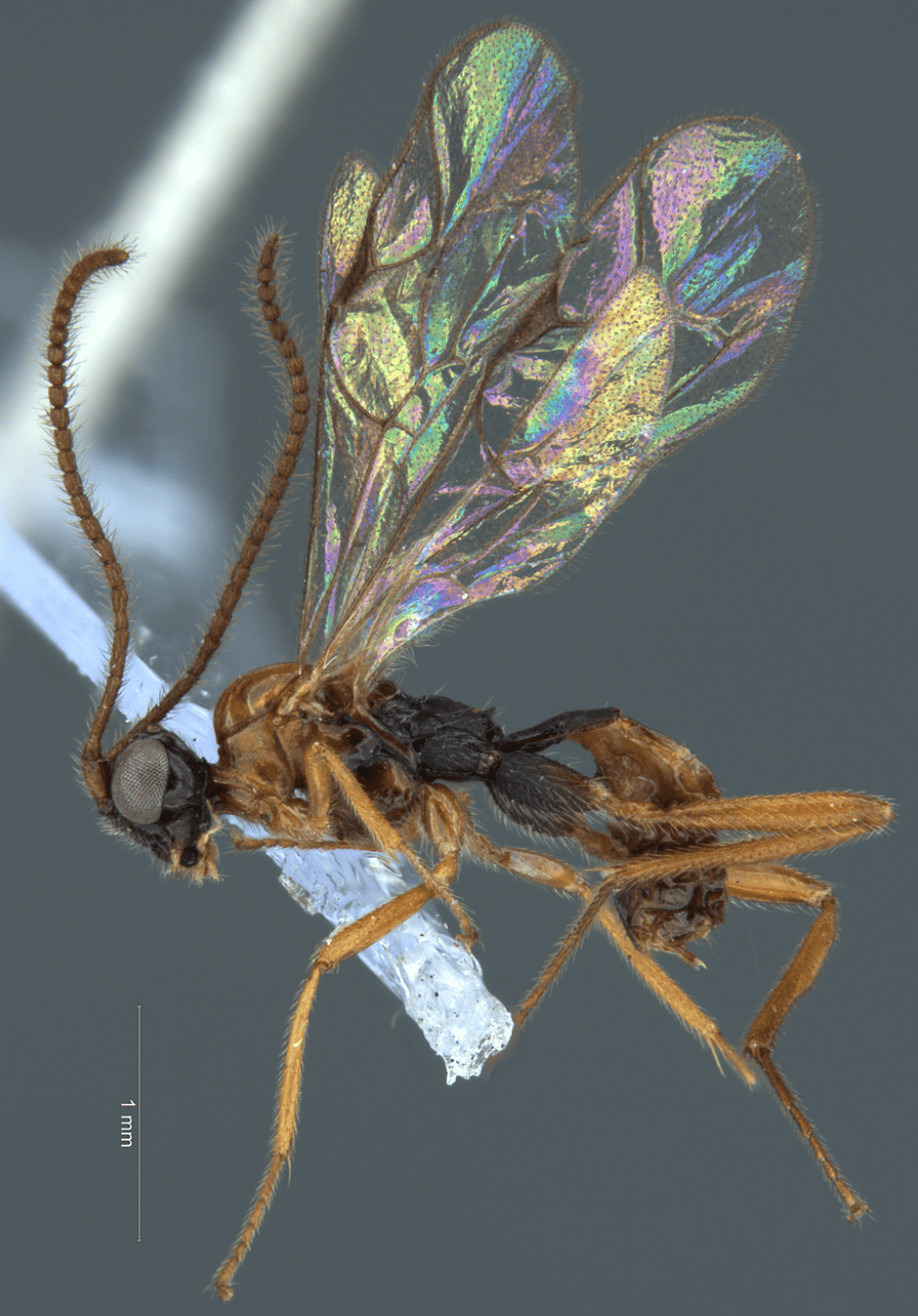}
        \par (j)
    \end{minipage}
    \begin{minipage}[t]{0.35\textwidth}
        \centering
        \includegraphics[width=\linewidth]{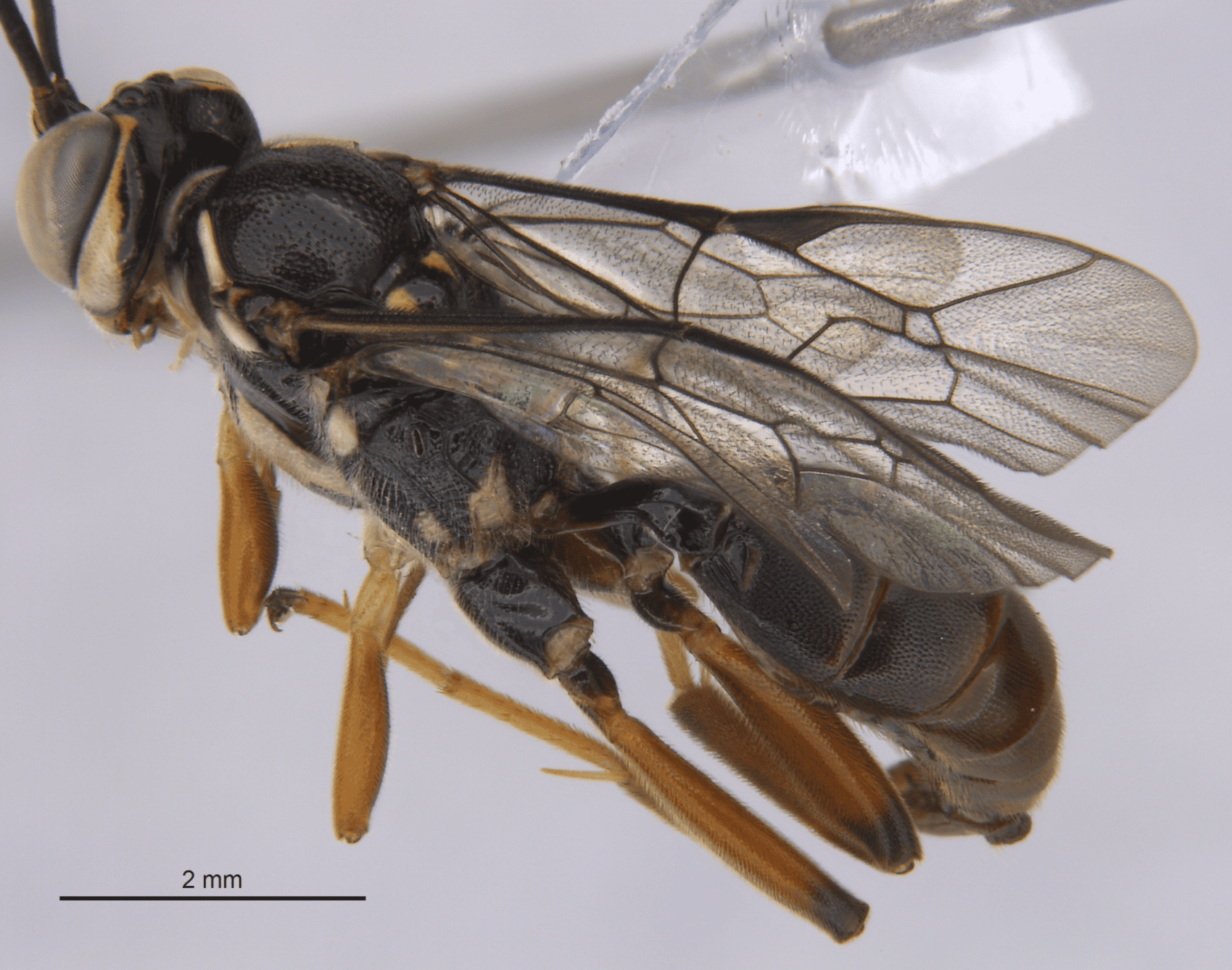}
        \par (k)
    \end{minipage}
    \hfill
    \begin{minipage}[t]{0.35\textwidth}
        \centering
        \includegraphics[width=\linewidth]{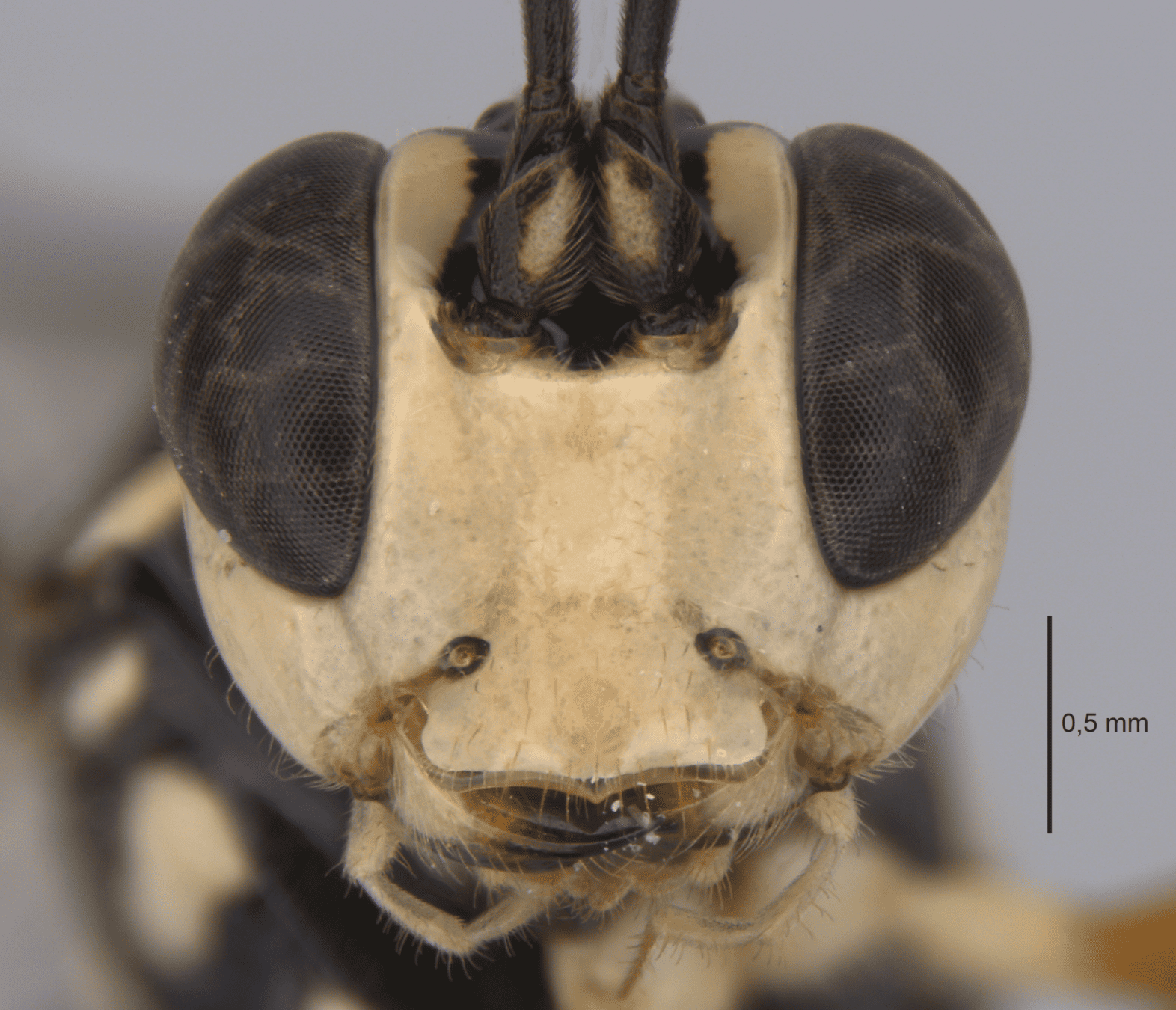}
        \par (l)
    \end{minipage}
    \caption{Examples of samples in the DAPWH dataset ~\cite{herrera_pinheiro_2026_18501018}. (a)-(f) Braconidae; (g)-(l) Ichneumonidae.}
    \label{dataset_samples_ich}
\end{figure*}

The Dataset of Parasitoid Wasps and Associated Hymenoptera (DAPWH) ~\cite{herrera_pinheiro_2026_18501018, Pinheiro11483097} comprises high-resolution images of Hymenoptera specimens, with a primary focus on the families Ichneumonidae and Braconidae from the DCBU taxonomic collection at UFSCar. Figure~\ref{fig_species} illustrates the photographic workflow from specimen collection. The dataset contains a total of 3,556 images, of which more than 40\% correspond to Ichneumonoidea wasps. Figure~\ref{dataset_samples_ich} shows some samples of these wasps.
% The biological material for this study, focusing on the Ichneumonoidea, was sourced from the DCBU taxonomic collection at UFSCar. Following specimen retrieval, morphological documentation was conducted using a Leica M205C stereomicroscope paired with a K5C digital camera. The acquisition process was managed via LAS X software, while the final high-depth-of-field composites were generated through digital image stacking in Helicon Focus. Figure~\ref{fig_species} illustrates the photographic workflow from specimen collection to final image processing.

% The Dataset of Parasitoid Wasps and Associated Hymenoptera (DAPWH) ~\cite{herrera_pinheiro_2026_18501018, Pinheiro11483097} comprises high-resolution images of Hymenoptera specimens, with a primary focus on the families Ichneumonidae and Braconidae. The dataset contains a total of 3,556 images, of which more than 40\% correspond to Ichneumonoidea wasps. Figure~\ref{dataset_samples_ich} shows some samples of these wasps.
% \begin{table}[h]
% \centering
% \caption{Distribution of images per family in DAPWH.}\label{tbl1}
% \begin{tabular}{|c|c|}
% \toprule
%         Family & Images \\ 
% \midrule
%         Ichneumonidae   & 786 \\
%         Braconidae      & 648 \\
%         Apidae          & 466 \\
%         Vespidae        & 460 \\
%         Megachilidae    & 298 \\
%         Chrysididae     & 244 \\
%         Andrenidae      & 244 \\
%         Pompilidae      & 190 \\
%         Bethylidae      & 94  \\
%         Halictidae      & 75  \\
%         Colletidae      & 51  \\
% \midrule
%         \textbf{Total} &\textbf{3,556} \\ 
% \bottomrule
% \end{tabular}
% \end{table}

\subsection{Model architecture}
For the automated identification of Ichneumonoidea, we selected the YOLOv12 ~\cite{tian2025yolo12} and YOLOv26 ~\cite{sapkota2026yolo26keyarchitecturalenhancements} architectures. These models represent the current state-of-the-art in object detection and classification, offering peak performance for complex biological datasets. Their selection was essential for processing the intricate morphological data found in the DAPWH dataset, experimental tests demonstrated that the YOLO framework achieved superior performance and faster training convergence compared to other evaluated architectures (e.g., ResNet, ViT, and VGG16) ~\cite{mastersthesis8a15de09,Pinheiro11483097}.

\subsection{Data splitting and imaging rescale}
For the model development phase, the dataset was partitioned into three distinct subsets to ensure robust training and unbiased evaluation. Following established methodological conventions ~\cite{raschka2020modelevaluationmodelselection}, we allocated 70\% of the total images for the training set, while the remaining 30\% was divided equally, with 15\% dedicated to validation during training and 15\% reserved as an independent test hold-out set. This distribution ensures that the final performance metrics represent the model's ability to generalize to unseen Ichneumonoidea specimens. The final partitioning of the dataset into training, validation, and test subsets is detailed in Table~\ref{tab:dataset_distribution}

\begin{table}[h]
\centering
\caption{Dataset distribution by family after splitting.}
\label{tab:dataset_distribution}
\begin{tabular}{|c|c|c|c|c|}
\toprule
Family & Train & Val & Test & Total \\ \midrule
Andrenidae      & 170            & 36           & 38            & 244            \\
Apidae          & 326            & 69           & 71            & 466            \\
Bethylidae      & 65             & 14           & 15            & 94             \\
Braconidae      & 453            & 97           & 98            & 648            \\
Chrysididae     & 170            & 36           & 38            & 244            \\
Colletidae      & 35             & 7            & 9             & 51             \\
Halictidae      & 52             & 11           & 12            & 75             \\
Ichneumonidae   & 550            & 117          & 119           & 786            \\
Megachilidae    & 208            & 44           & 46            & 298            \\
Pompilidae      & 133            & 28           & 29            & 190            \\
Vespidae        & 322            & 69           & 69            & 460            \\ \midrule
\textbf{Total} & \textbf{2,484} & \textbf{528} & \textbf{544} & \textbf{3,556} \\ \bottomrule
\end{tabular}
\end{table}
Given the high-fidelity nature of the original stacked images acquired with the Leica M205C system, spatial downsampling was required to align with the neural network's computational constraints. All images were rescaled to a fixed input dimension of $512~\times~512$ pixels for YOLO training. 

\subsection{Training and evaluation}
The training and evaluation of the models were performed on a high-performance workstation running Linux. The hardware configuration consisted of an AMD Ryzen 9 7900 CPU, 64GB of DDR5 RAM, and an NVIDIA RTX 4090 GPU with 24GB of VRAM with CUDA 13.1. The source code are publicly available in \href{https://github.com/joaomh/identification-of-Ichneumonoidea-waps-YOLO-2026}{GitHub}.

To quantify the classification performance of the developed models, we employed a suite of standard evaluation metrics: Accuracy, Precision, Recall, and the F1-score. These indicators are widely recognized as benchmarks for both image classification and object detection tasks~\cite{lin2015microsoftcococommonobjects}.
%defined by Eqs.~\ref {accuracy_equation}, \ref {precision_equation}, \ref {recall_equation}, and \ref {eq:f1_score}, respectively.%

% \begin{equation}
% \text{Accuracy} = \frac{TP + TN}{TP + TN + FP + FN},
% \label{accuracy_equation}
% \end{equation}
%   \begin{equation}
%         \text{Precision} = \frac{TP}{TP + FP}.
%         \label{precision_equation}
%     \end{equation}
%    \begin{equation}
%         \text{Recall} = \frac{TP}{TP + FN}.
%         \label{recall_equation}
%     \end{equation}
%     \begin{equation}
%         F_1 = \frac{2 \cdot \text{Precision} \cdot \text{Recall}}{\text{Precision} + \text{Recall}}.
%         \label{eq:f1_score}
%     \end{equation}
% where $TP$, $TN$, $FP$, and $FN$ represent true positives, true negatives, false positives, and false negatives, respectively.

\subsection{Model interpretability}
While quantitative metrics such as accuracy, precision, recall, and F1-score are statistical measures of performance, they do not reveal the neural network's decision-making process. To ensure the taxonomic validity of the model's predictions, we employed Explainable AI (XAI) techniques, specifically High-Resolution Class Activation Mapping (HiRes-CAM) ~\cite{draelos2021usehirescaminsteadgradcam}. HiRes-CAM computes element-wise importance scores to produce visualization maps that are strictly faithful to the model's computations. This higher spatial precision allows us to verify whether the model is focusing on relevant morphological diagnostic traits, such as specific wing venation patterns, rather than learning spurious correlations from the background.
% \subsection{Research workflow}
% Figure~\ref{Research workflow} shows the overview of the proposed explainable identification framework. High-resolution images of Ichneumonidae specimens are provided as input to the YOLOv26 model. The network extracts hierarchical feature representations that capture discriminative patterns across convolutional layers. Finally, HiResCAM is applied to generate class-discriminative activation maps that highlight biologically relevant regions, such as wing venation and thoracic structures, to support transparent and interpretable predictions.

% \begin{figure*}
%     \centering
%         \includegraphics[width=0.98\linewidth]{figs/flow_images_yolo_cam.png}
%     \caption{Research workflow.}
%     \label{Research workflow}
% \end{figure*}
\section{Results and discussion}
\label{sec:results}
\subsection{Model performance}

The performance evaluation conducted on the DAPWH test set indicates that both architectures achieve high levels of taxonomic discrimination for the Ichneumonoidea superfamily. As summarized in Table~\ref{tab:model_comparison}, the YOLOv26 model demonstrated superior performance across all evaluated metrics. 

Specifically, YOLOv26 achieved a Top-1 Accuracy of 96.14\%, representing a significant improvement over the 94.85\% attained by the YOLOv12 variant. Regarding the model's reliability in identifying complex morphological features, YOLOv26 reached a Precision of 93.43\% and a robust Recall of 97.04\%. The resulting $F_1$-score of 95.20\% further confirms the model's effectiveness in balancing false positives and negatives. 

\begin{table}[h]
\centering
\caption{Performance Comparison of YOLO Classification Models on the DAPWH Test Set.}
\label{tab:model_comparison}
\begin{tabular}{|c|c|c|c|c|}
\toprule
Model & Accuracy & Precision & Recall & $F_{1}$ \\
YOLOV12   & 0.9485 & 0.9132 & 0.9429  & 0.9278  \\
YOLOV26 & 0.9614 & 0.9343 & 0.9704  & 0.9520  \\ 
\bottomrule
\end{tabular}
\end{table}
Both models exhibited stable convergence over the 150 training epochs, with a rapid reduction in training loss during the initial iterations followed by gradual stabilization. For YOLOv12, the training loss decreased sharply within the first 20–30 epochs and asymptotically approached near-zero values, while the validation loss stabilized around ~0.20 after early fluctuations. The Top-1 accuracy increased consistently, surpassing 0.95 in later epochs, whereas Top-5 accuracy rapidly saturated, remaining close to 1.00 throughout most of the training process. These trends indicate efficient feature learning and strong generalization capacity without evident signs of overfitting.

Similarly, YOLOv26 (Figure 4b) demonstrated fast convergence and improved stability during validation. The validation loss exhibited slightly lower variance than YOLOv12 and converged to marginally lower values. Top-1 accuracy steadily improved to approximately 0.97 in the final epochs, indicating robust ranking performance. Overall, both architectures achieved high classification accuracy; however, YOLOv26 presented smoother validation behavior and slightly superior generalization performance.

\begin{figure}[ht]
    \centering
        \includegraphics[width=0.8\linewidth]{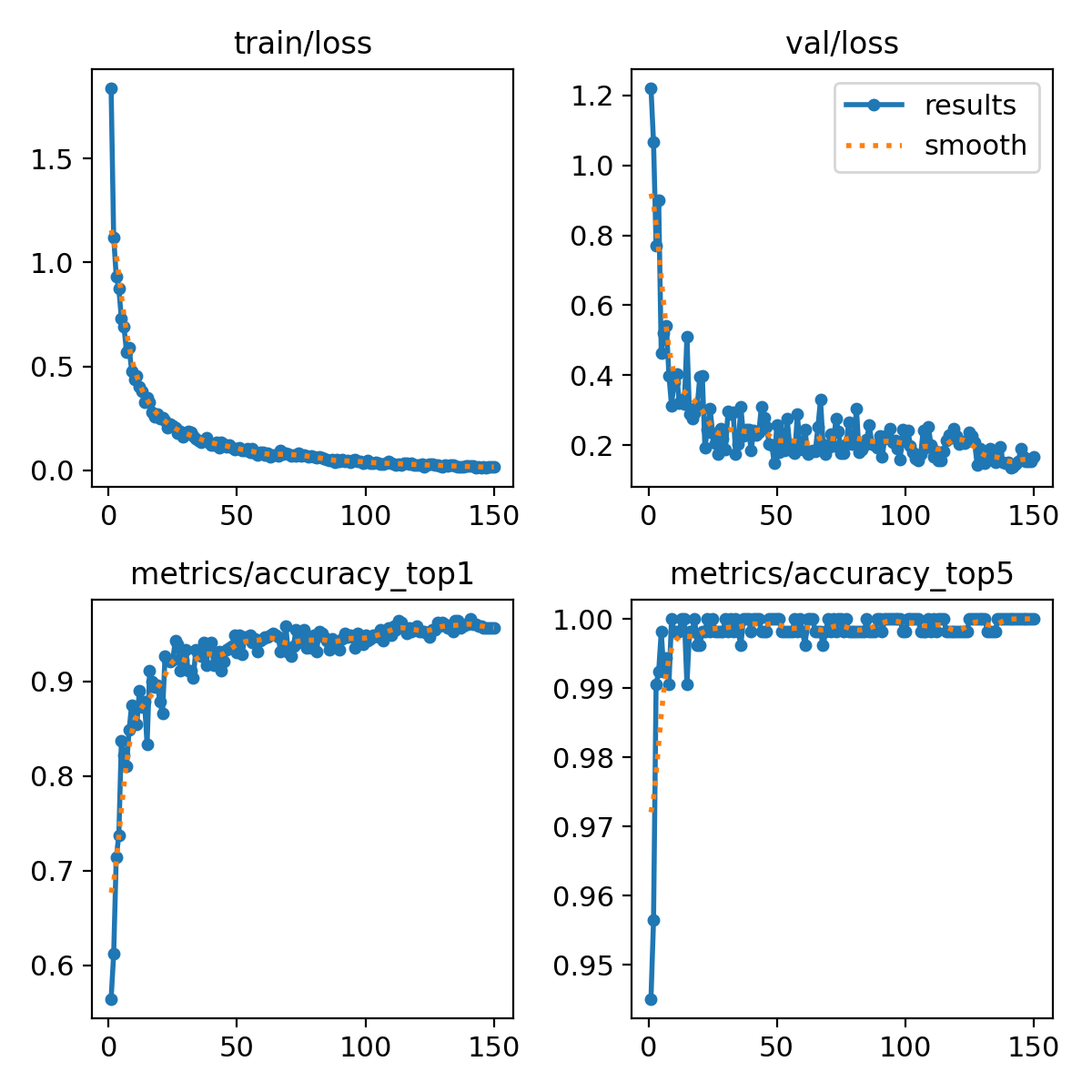}
    \caption{Training and validation performance metrics of the YOLOv12 architecture over 150 epochs. The subplots illustrate the progression of training loss, validation loss, Top-1 accuracy, and Top-5 accuracy. The continuous decrease and subsequent stabilization of both training and validation losses, without significant divergence between the curves, provide empirical evidence against overfitting. Furthermore, both Top-1 and Top-5 accuracy metrics demonstrate rapid early convergence, stabilizing at high plateau values and indicating robust learning dynamics across the dataset.}
    \label{dataset_val_curves_v12}
\end{figure}

\begin{figure}[ht]
    \centering
        \includegraphics[width=0.8\linewidth]{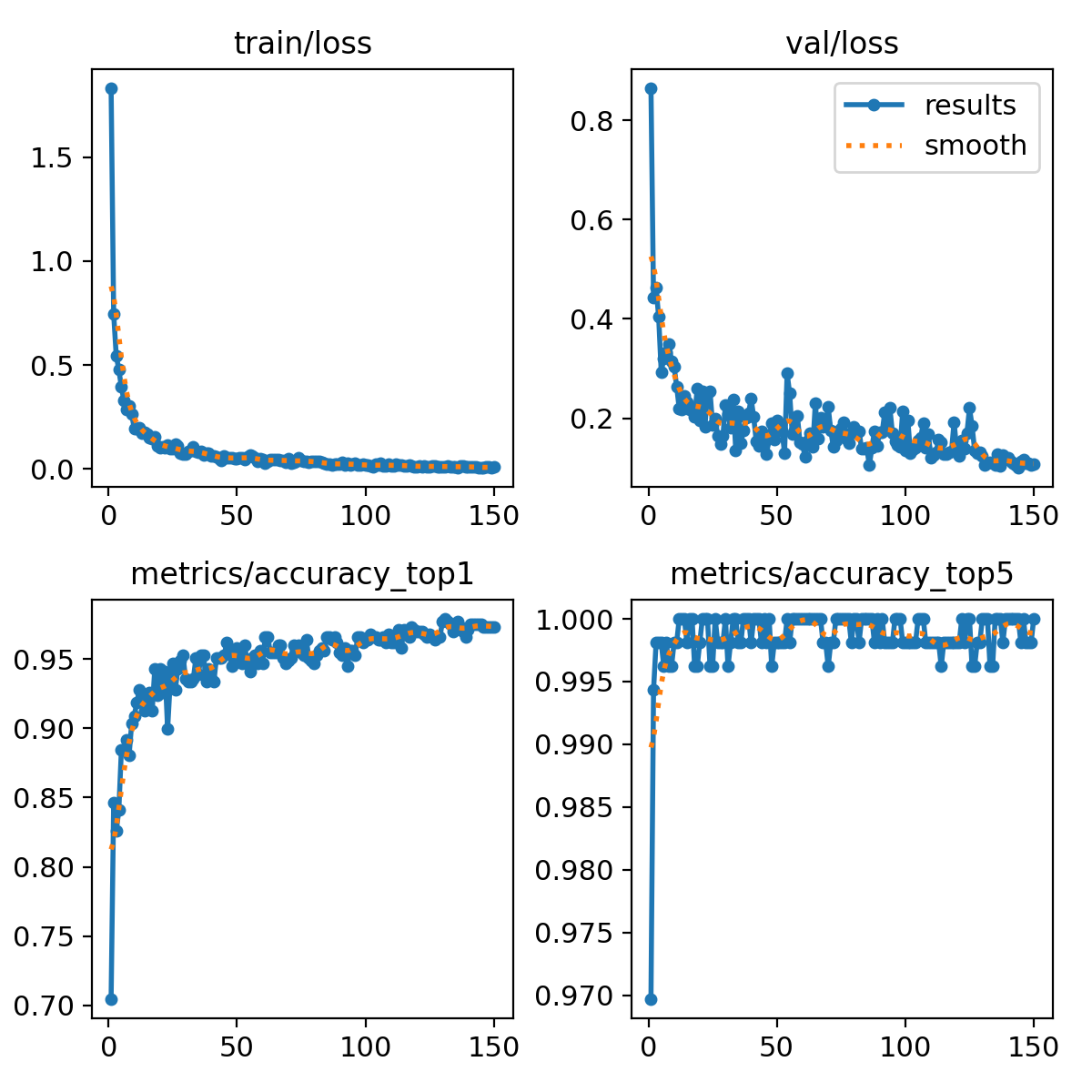}
    \caption{Training and validation performance metrics of the YOLOv26 architecture over 150 epochs. Similar to the YOLOv12 evaluation, the validation loss steadily decreases in tandem with the training loss, indicating strong generalization capabilities to unseen data and an absence of overfitting.}
    \label{dataset_val_curves_v26}
\end{figure}

\begin{figure*}
    \centering
    \begin{minipage}[t]{0.8\textwidth}
        \centering
        \includegraphics[width=\linewidth]{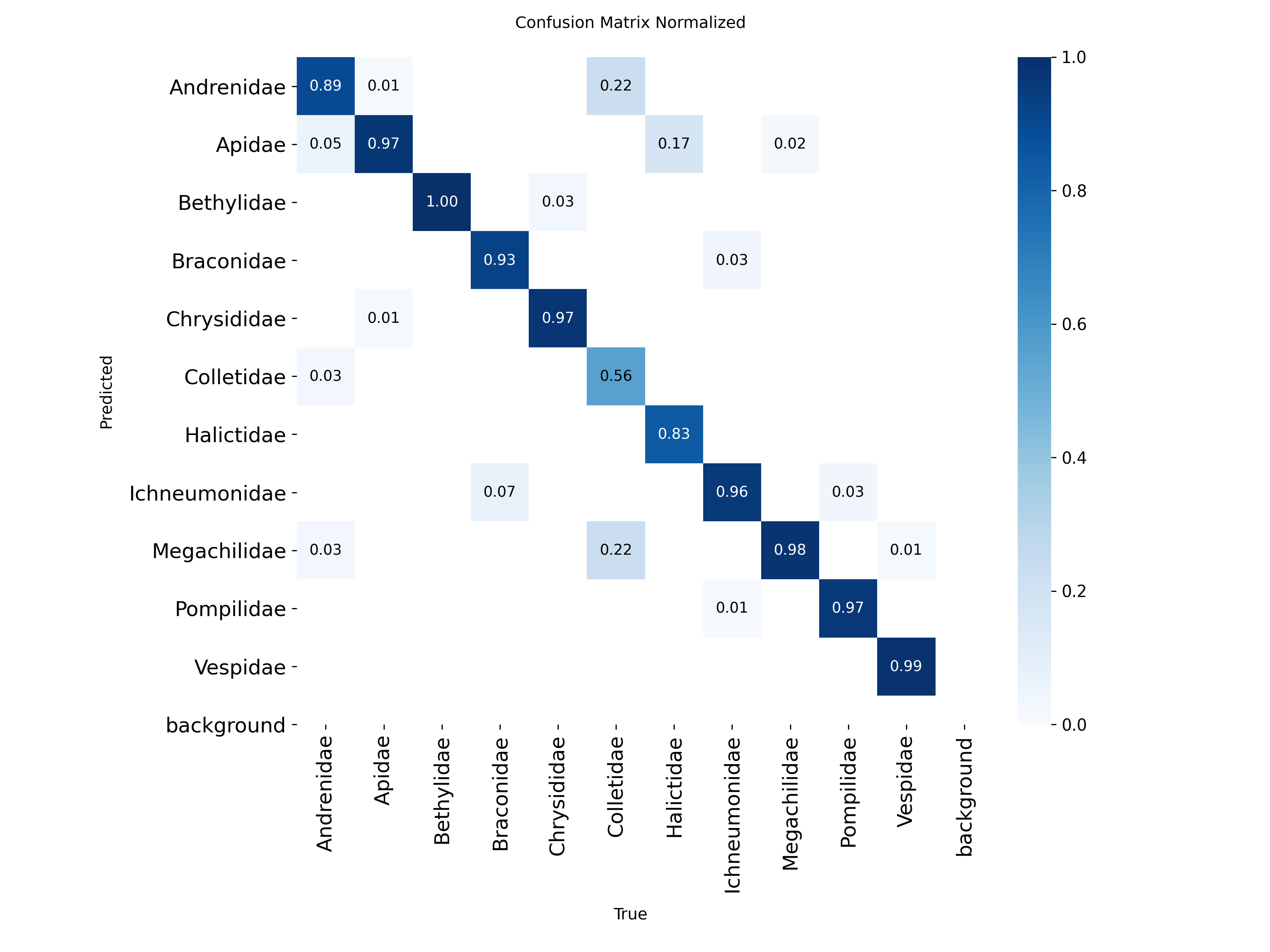}
        \par (a)
    \end{minipage}
    \vspace{0.1cm} % Add some vertical spacing between rows
    \begin{minipage}[t]{0.8\textwidth}
        \centering
        \includegraphics[width=\linewidth]{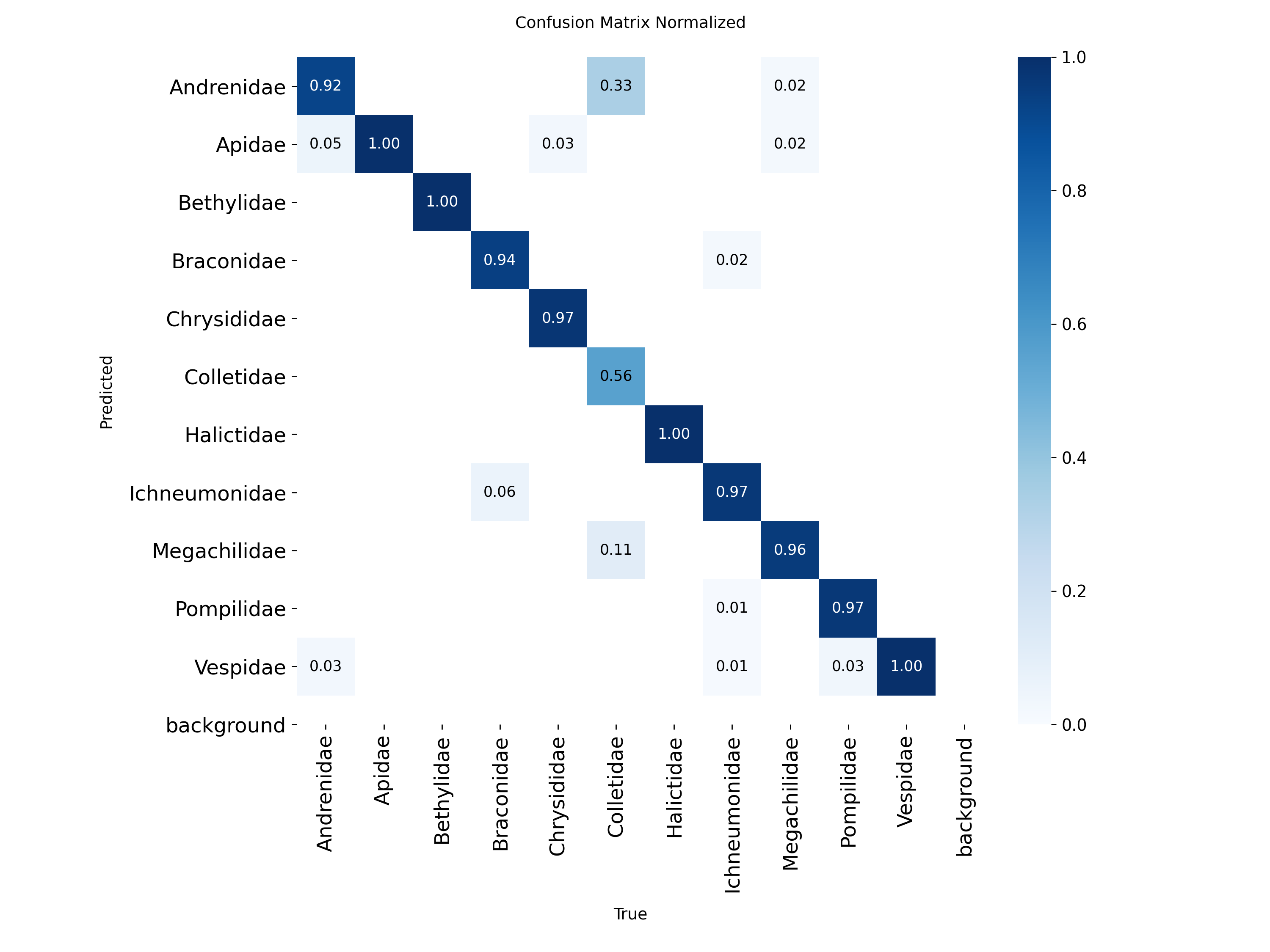}
        \par (b)
    \end{minipage}
    \caption{Confusion matrix normalized. (a) YOLOv12; (b) YOLOv26.}
    \label{matrix:yolov}
\end{figure*}

The normalized confusion matrices (Figure ~\ref{matrix:yolov}) demonstrate strong class-level discrimination for both models, with dominant diagonal values indicating high per-family accuracy. For YOLOv12, most families achieved correct identification rates above 93\%, including Ichneumonidae and Braconidae.
% Moderate confusion was observed for Colletidae and Halictidae, suggesting greater morphological similarity or class imbalance effects. Limited cross-family misclassification occurred primarily between taxonomically related groups, such as Andrenidae and Halictidae, and between Megachilidae and Colletidae.

YOLOv26 showed improved overall discrimination. Ichneumonidae achieved 97\% accuracy, while Braconidae remained above 94\%. The concentration of high diagonal values and the reduction of off-diagonal errors indicate enhanced generalization and more consistent inter-family boundary learning in YOLOv26.

\subsection{Model interpretability}
The qualitative analysis of the learned representations and attention maps provides further insight into the model's internal decision-making process. The feature activation maps extracted from intermediate convolutional layers reveal that the network progressively encodes discriminative morphological patterns, emphasizing structural contours while suppressing background information. The diversity of activation responses across channels indicates hierarchical feature abstraction, ranging from low-level edge detection to higher-level morphological descriptors.

\begin{figure*}
    \centering
    \begin{minipage}[t]{0.9\textwidth}
        \centering
        \includegraphics[width=\linewidth]{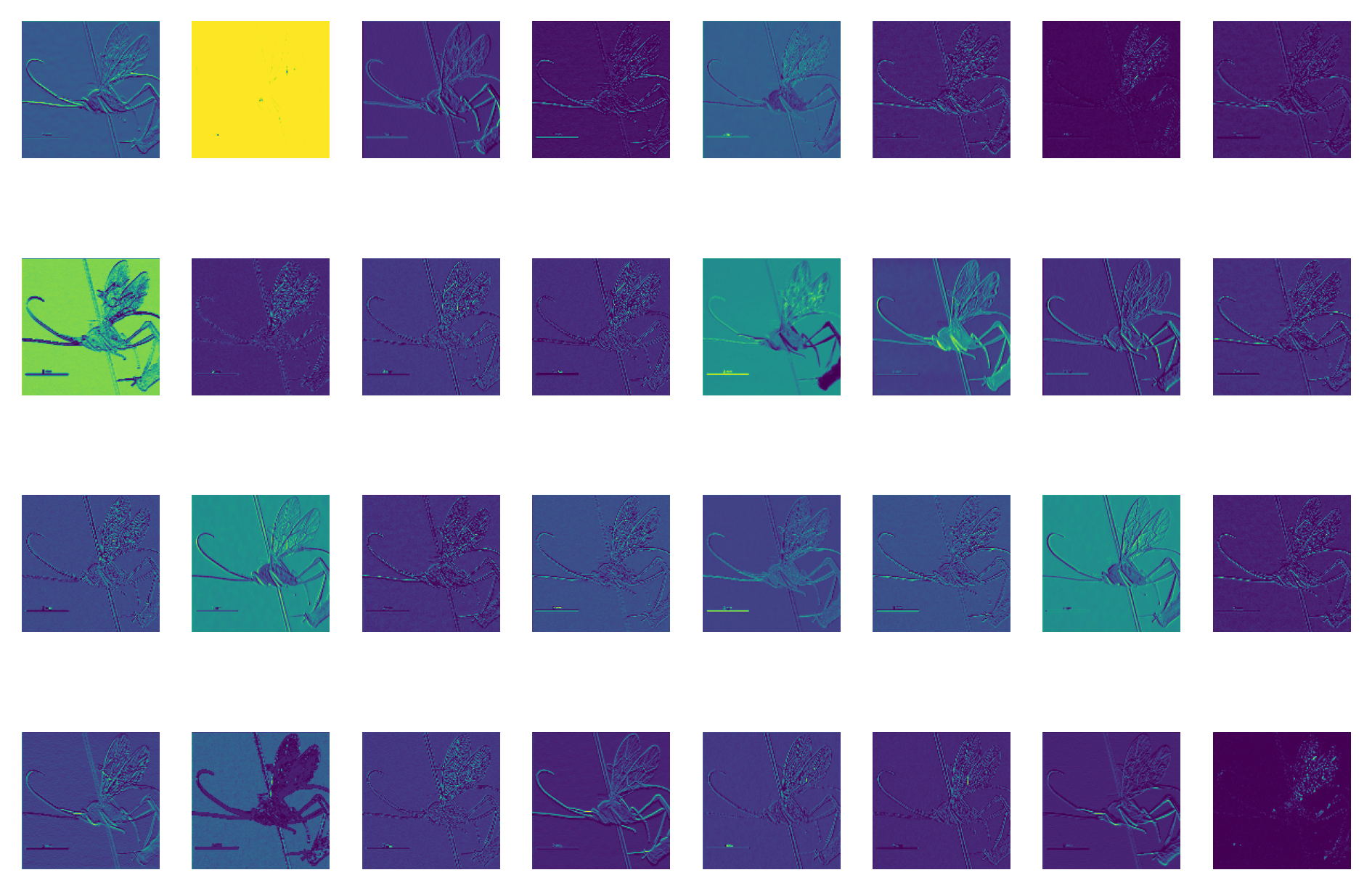}
        \par (a)
    \end{minipage}
    \vspace{0.1cm} % Add some vertical spacing between rows
    \begin{minipage}[t]{0.9\textwidth}
        \centering
        \includegraphics[width=\linewidth]{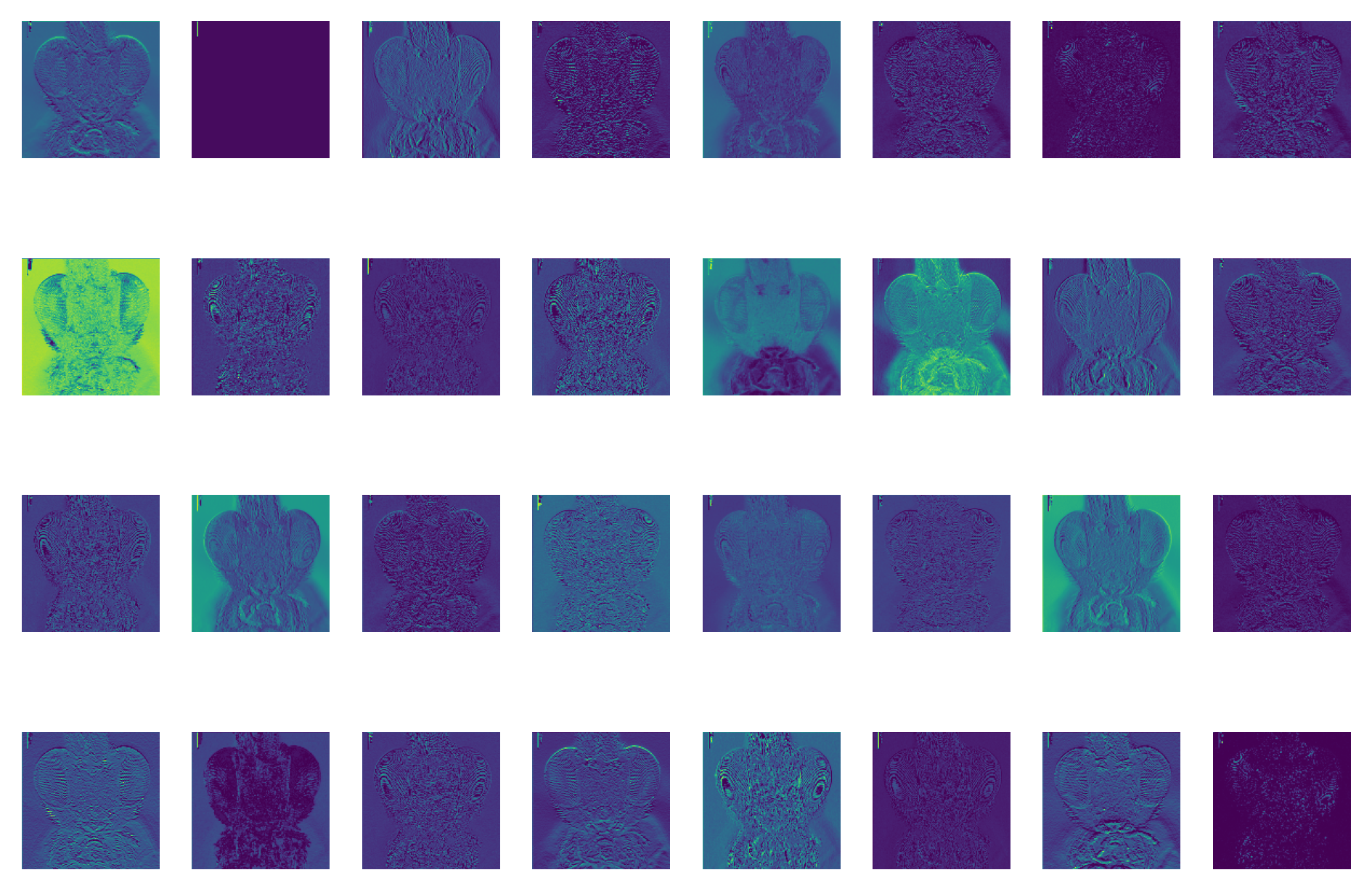}
        \par (b)
    \end{minipage}
    \caption{Representative feature maps samples extracted from intermediate convolutional layers for YOLOv26, illustrating the hierarchical encoding of morphological structures and texture patterns for Ichneumonidae. (a) Habitus lateral; (b) Head frontal}
    \label{conv_model_ich_correct}
\end{figure*}

\begin{figure*}
    \centering
    \begin{minipage}[t]{0.9\textwidth}
        \centering
        \includegraphics[width=\linewidth]{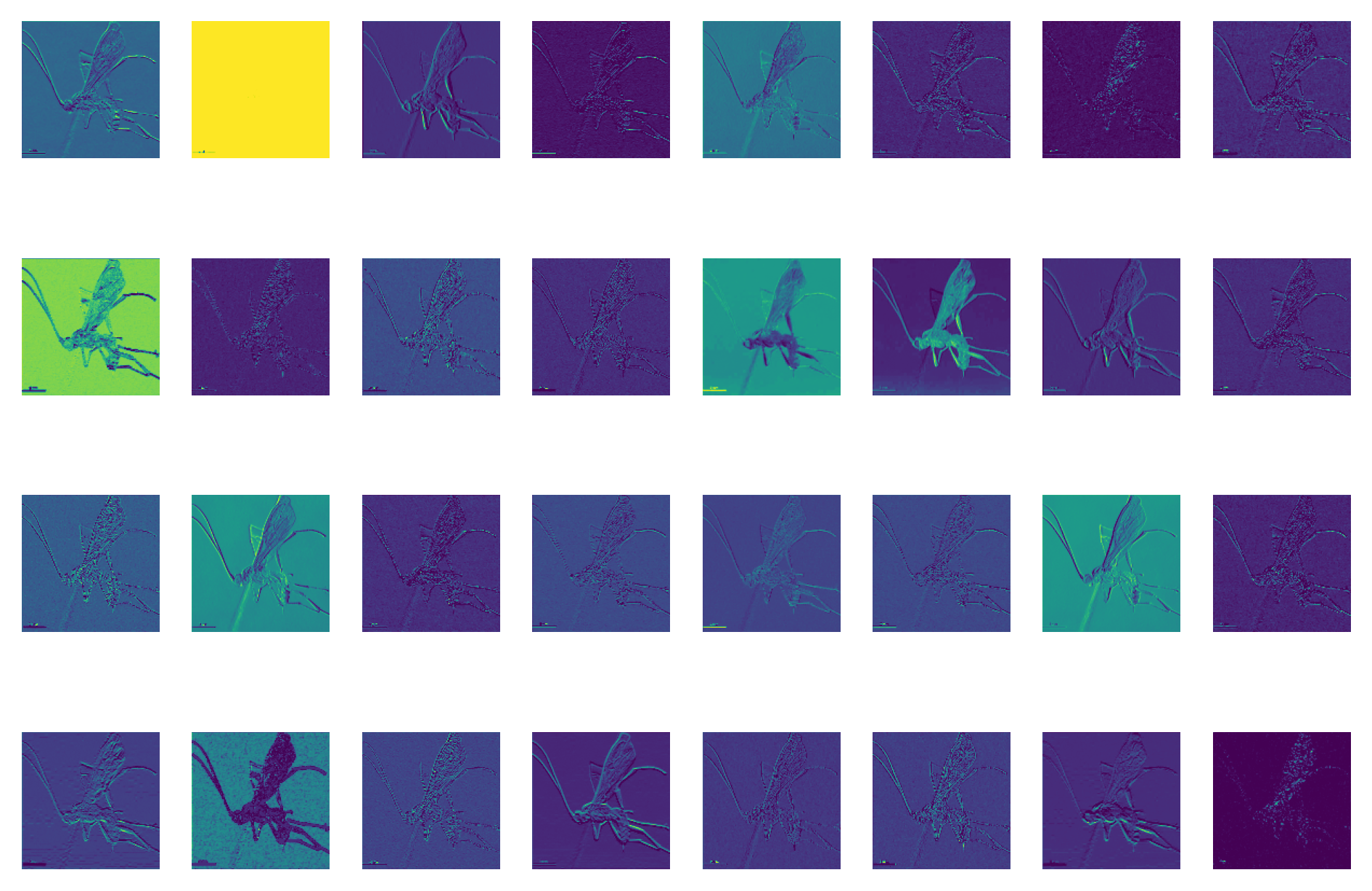}
        \par (a)
    \end{minipage}
    \vspace{0.1cm} % Add some vertical spacing between rows
    \begin{minipage}[t]{0.9\textwidth}
        \centering
        \includegraphics[width=\linewidth]{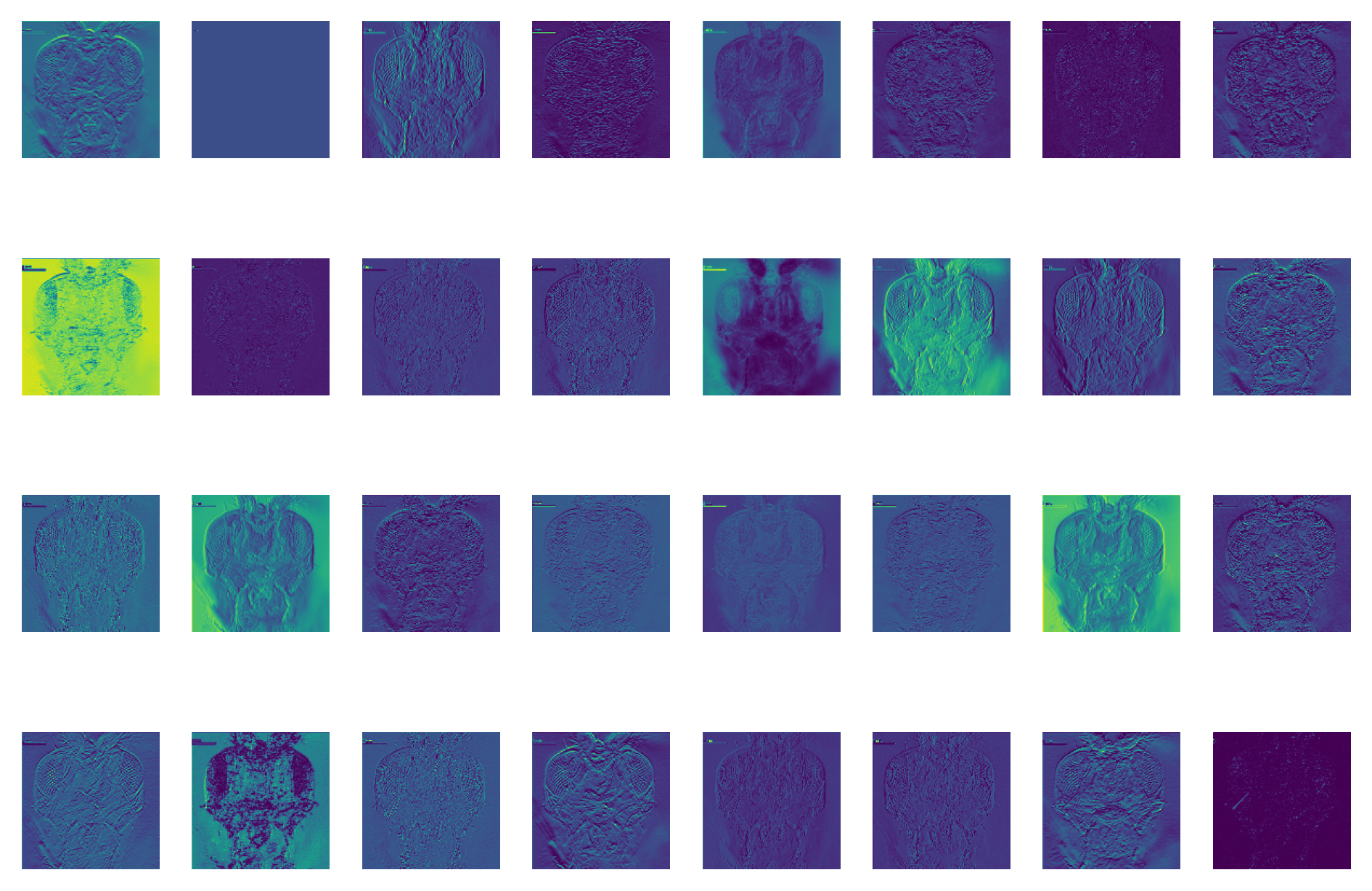}
        \par (b)
    \end{minipage}
    \caption{Representative feature maps samples extracted from intermediate convolutional layers for YOLOv26, illustrating the hierarchical encoding of morphological structures and texture patterns for Braconidae. (a) Habitus lateral; (b) Head frontal}
    \label{conv_model_braco_correct}
\end{figure*}

In Figure ~\ref{conv_model_ich_correct}, the visualizations demonstrate that the model emphasizes critical structural contours, such as the wing segmentation for the family Ichneumonidae. For the family Braconidae, the visualization of intermediate convolutional layers demonstrates that the model also effectively suppresses background noise. As shown in Figure ~\ref{conv_model_braco_correct}, the hierarchical encoding process prioritizes diagnostically relevant anatomical regions, such as the metasomal segmentation.
By capturing these multi-scale features, the convolutional layers enable the model to achieve high per-family accuracies of 97\% for Ichneumonidae and 94\% for Braconidae, as shown in the normalized confusion matrix (Figure ~\ref{matrix:yolov}).

\subsection{Ichneumonidae}
\begin{figure}[H]
    \centering
        \includegraphics[width=0.8\linewidth]{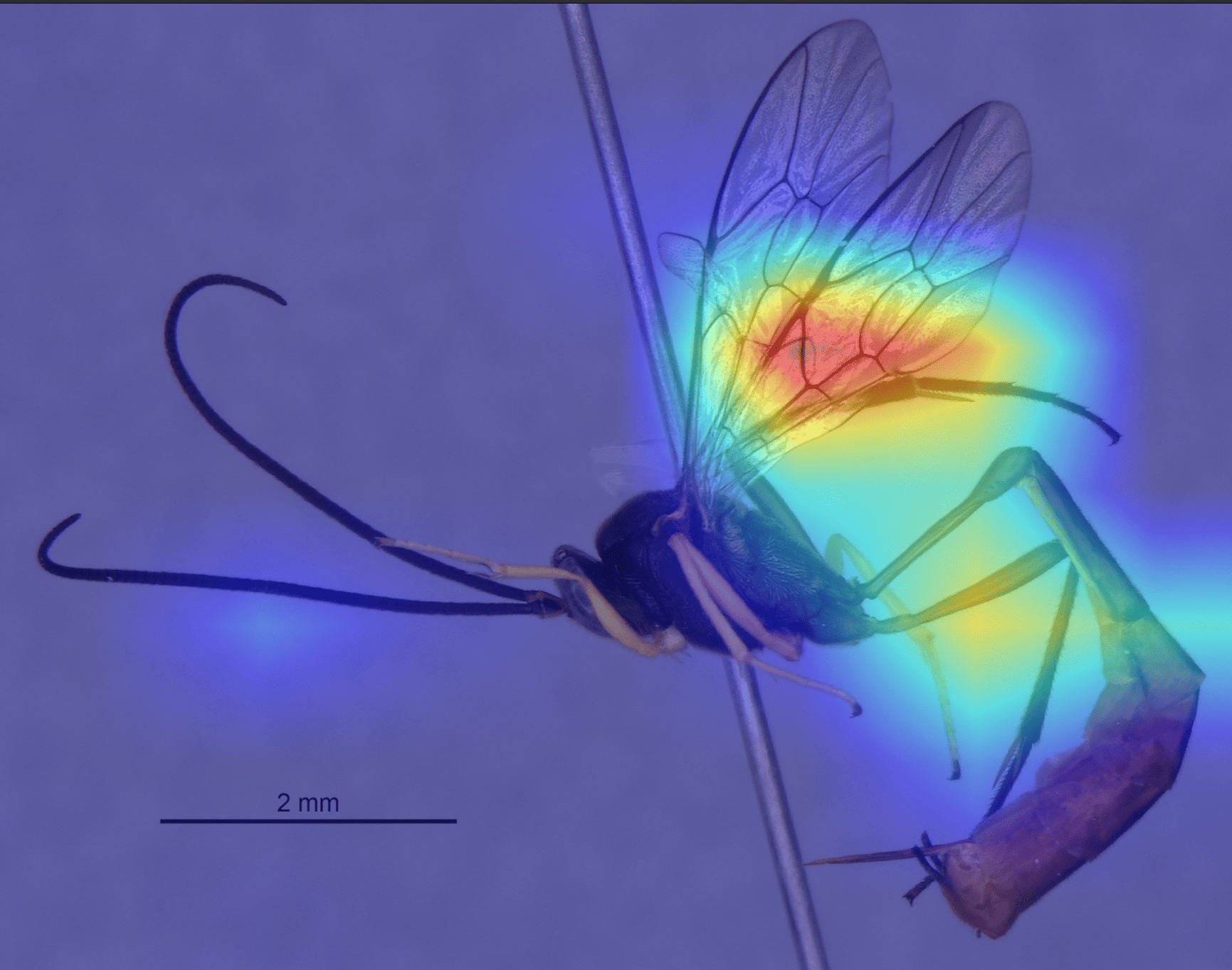}
	\caption{HiResCAM visualizations for Ichneumonidae. The heatmaps demonstrate that the YOLOv26 architecture prioritizes wing venation patterns, notably the discosubmarginal cell, aligning with established entomological keys.}
	\label{ich_correct_lateral_keys}
\end{figure}

\begin{figure}[H]
    \centering
        \includegraphics[width=0.8\linewidth]{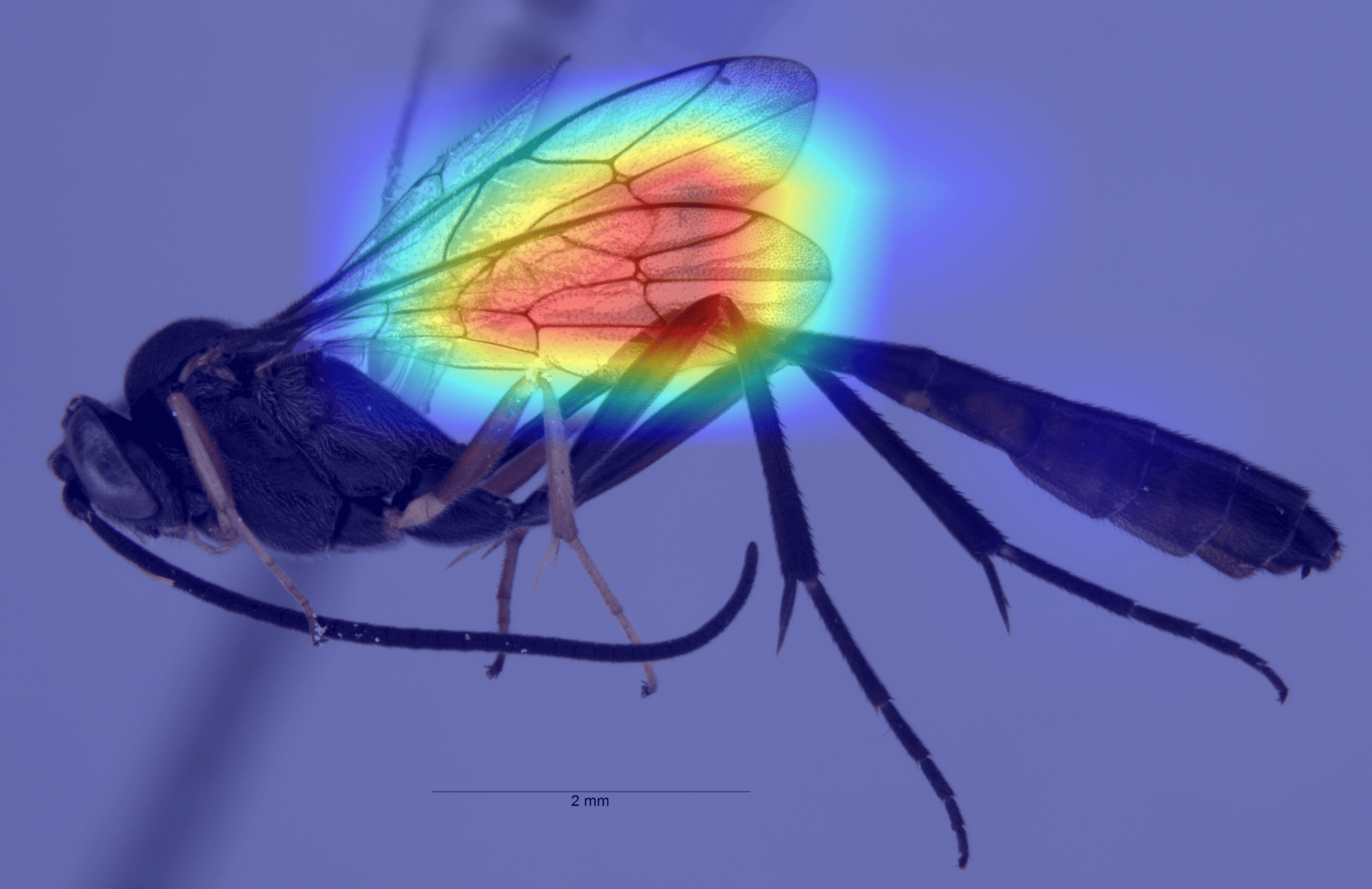}
	\caption{HiResCAM visualizations for Ichneumonidae. The heatmaps demonstrate that the YOLOv26 architecture prioritizes wing venation patterns, particularly the second recurrent vein (2m-cu), aligning with established entomological keys.}
	\label{ich_correct_lateral_keys_2}
\end{figure}

\begin{figure}[H]
    \centering
        \includegraphics[width=0.8\linewidth]{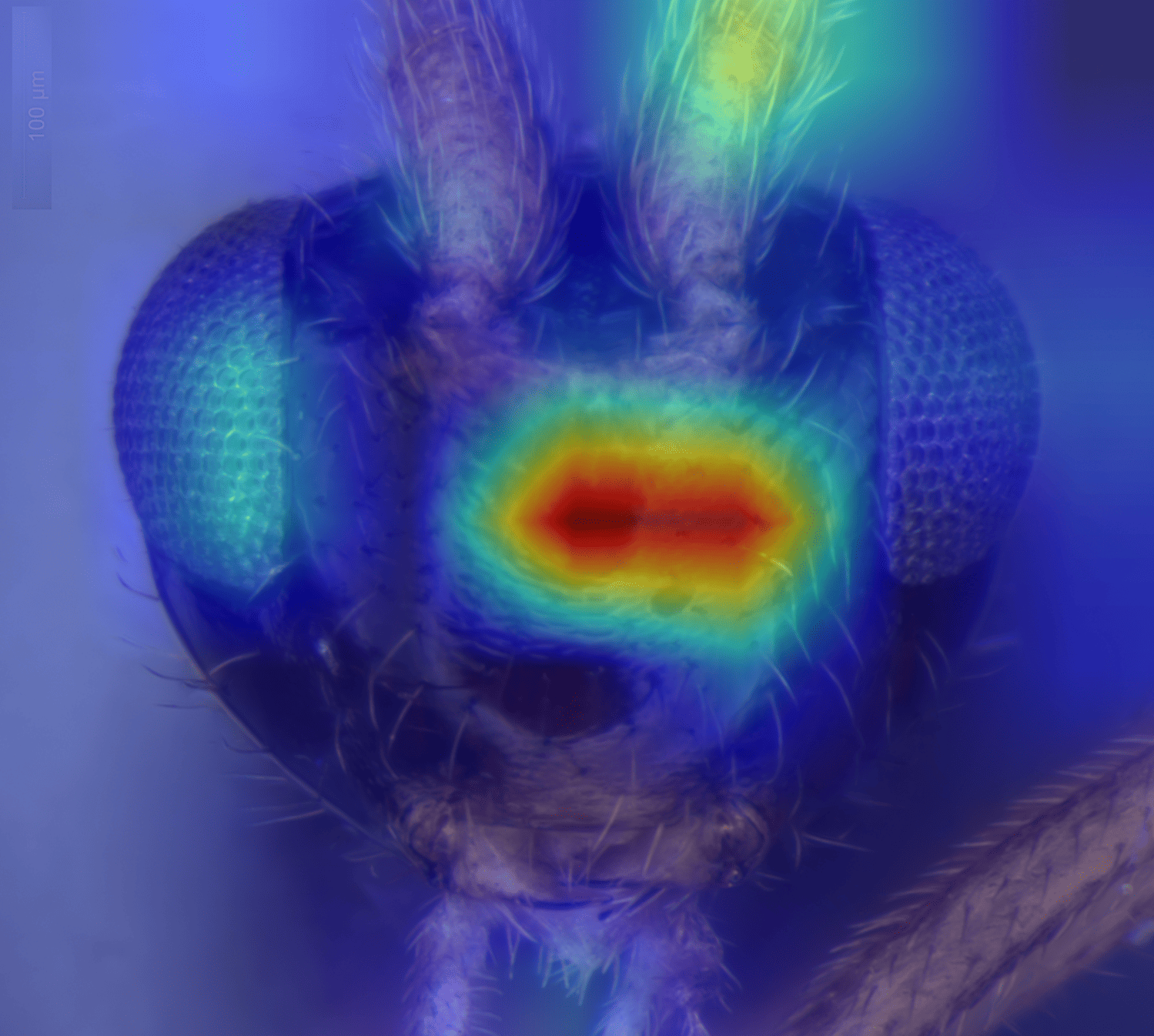}
	\caption{HiResCAM visualizations for Ichneumonidae, the heatmaps demonstrate that the YOLOv26 architecture prioritizes convex facial aligning with established entomological keys.}
	\label{ich_correct_frontal_keys}
\end{figure}

For the identification of Ichneumonidae, two situations were observed. In the first, the model likely relied on traditional morphological characters used to distinguish the family, particularly those of the fore wing (Figure \ref{ich_correct_lateral_keys} and Figure \ref{ich_correct_lateral_keys_2}). For example, the presence of the fore wing vein 2m-cu is a crucial character for identifying Ichneumonidae. This corresponds to step 2 in the key for the separation of British and Irish Braconidae and Ichneumonidae (Broad et al., 2018). Another important wing character is the absence of vein RS+M forming the discosubmarginal cell, which is used in step 3 of the same key. Additionally, facial features were also captured, such as the convex face typical of Ichneumonidae (Figure \ref{ich_correct_frontal_keys}).

\begin{figure}[H]
    \centering
        \includegraphics[width=0.8\linewidth]{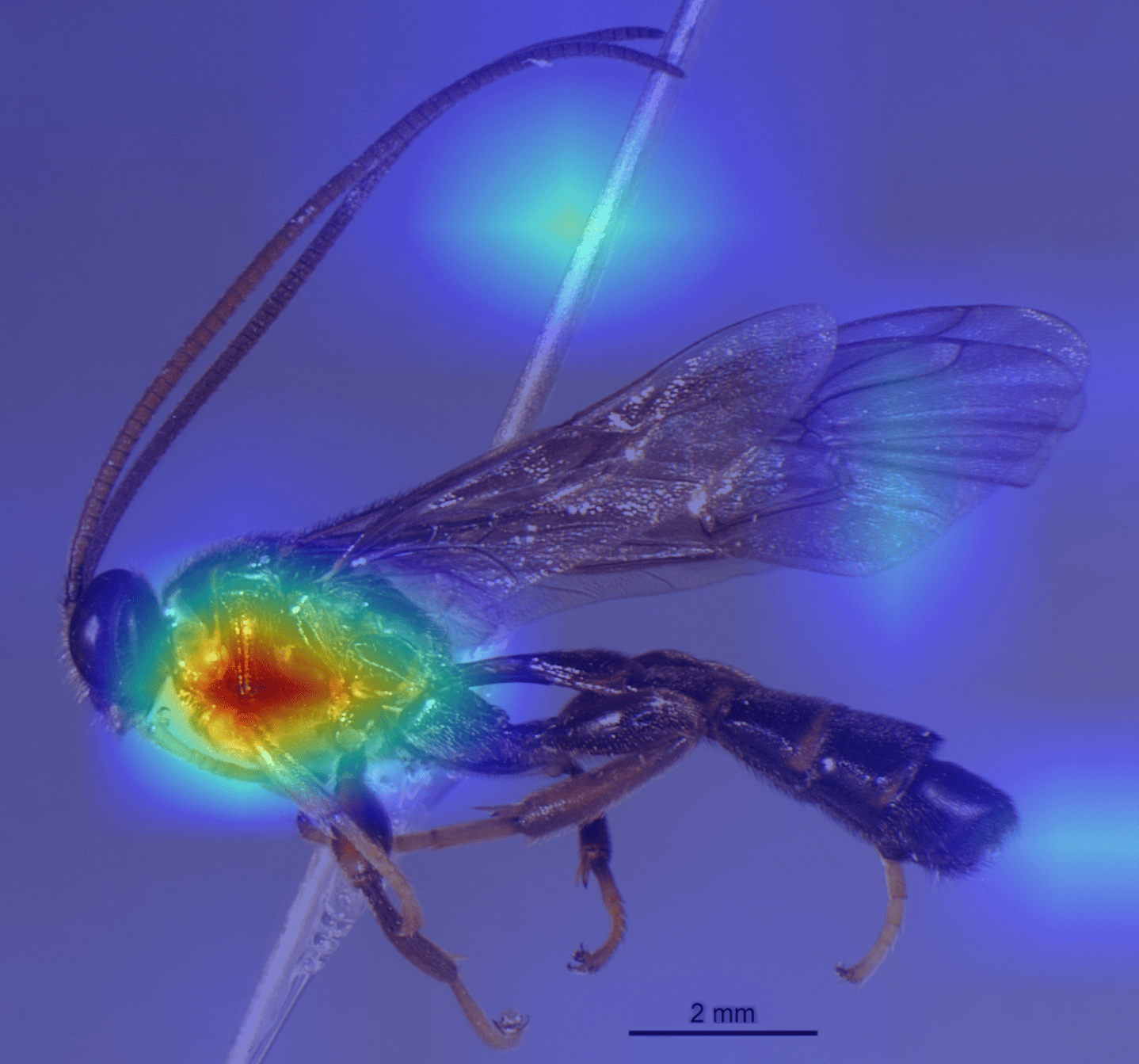}
	\caption{HiResCAM visualizations for Ichneumonidae reveal that the YOLOv26 architecture identifies and prioritizes alternative morphological structures beyond those traditionally emphasized in dichotomous taxonomic keys.}
	\label{ich_correct_lateral}
\end{figure}

In the second situation, however, the model appeared to rely on non-traditional diagnostic characteristics (Figure \ref{ich_correct_lateral}). Instead of focusing on explicit structural features commonly used in taxonomy, the model based its decisions on broader morphological patterns or overall visual similarity. This behavior highlights an opportunity to explore non-conventional or underemphasized characters that, while not formally incorporated into identification keys, could contain diagnostically significant information.

\subsection{Braconidae}

\begin{figure}[H]
    \centering
        \includegraphics[width=0.7\linewidth]{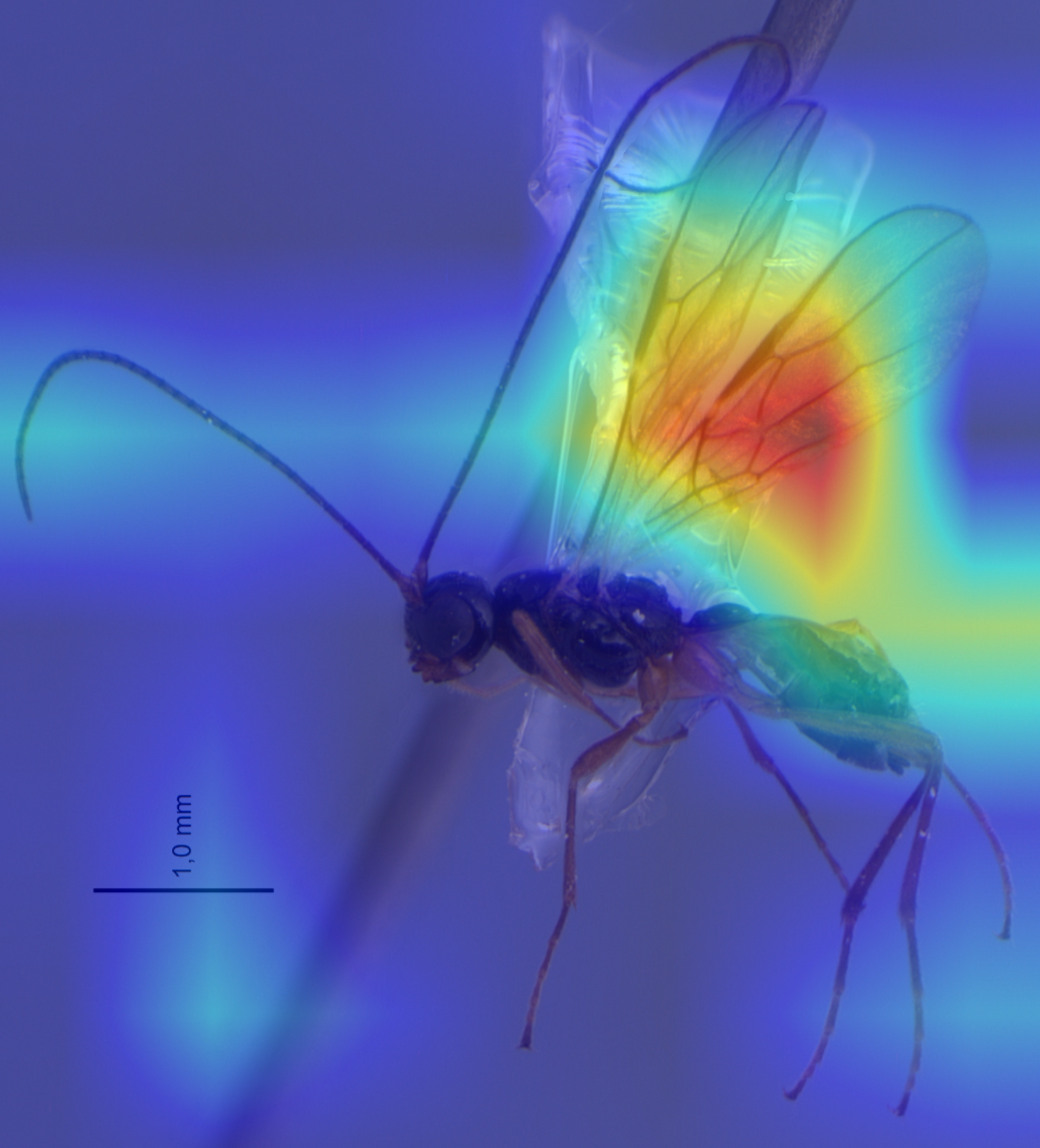}
	\caption{HiResCAM visualizations for Braconidae reveal that the YOLOv26 architecture identifies and prioritizes the absence of areolet and 2m-cu aligning with established entomological keys.}
	\label{braco_correct_lateral_2}
\end{figure}

The family Braconidae is characterized by several distinct morphological features successfully captured by the YOLOv26 architecture. The most reliable taxonomic distinction for this family is found in the wing venation ~\cite{Sharkey_key_2023}. Unlike Ichneumonids, Braconids almost invariably lack the second recurrent vein (2m-cu) and the areolet in the forewing (Figure \ref{braco_correct_lateral_2}).

\begin{figure}[H]
    \centering
        \includegraphics[width=0.8\linewidth]{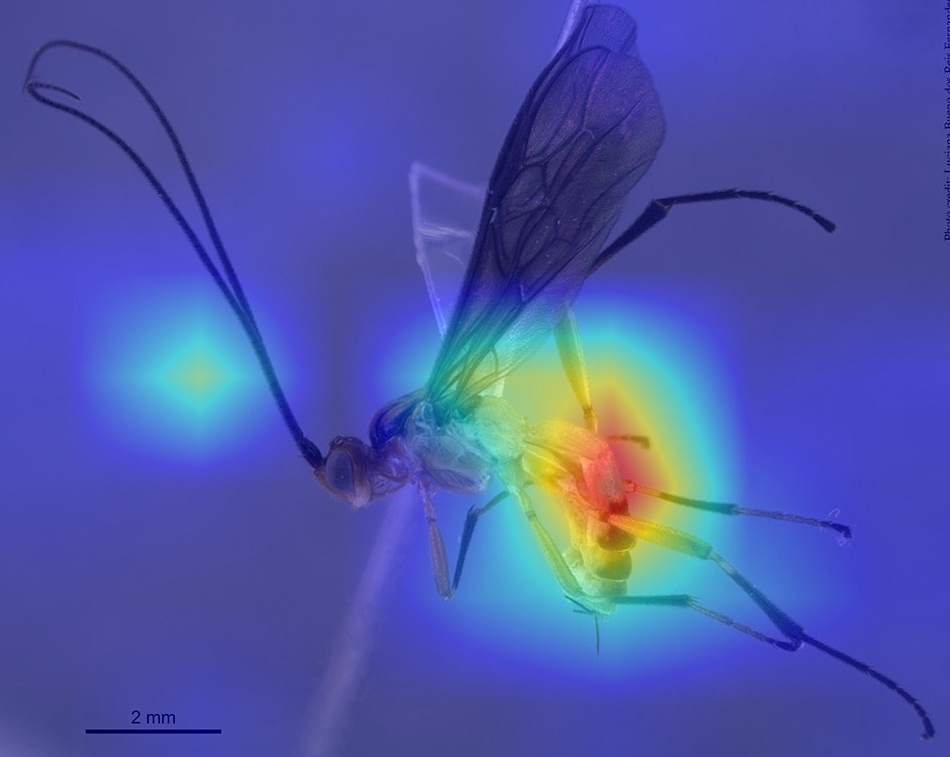}
	\caption{HiResCAM visualizations for Braconidae reveal that the YOLOv26 architecture identifies and prioritizes the fused metasomal aligning with established entomological keys.}
	\label{braco_correct_lateral}
\end{figure}

\begin{figure}[H]
    \centering
        \includegraphics[width=0.85\linewidth]{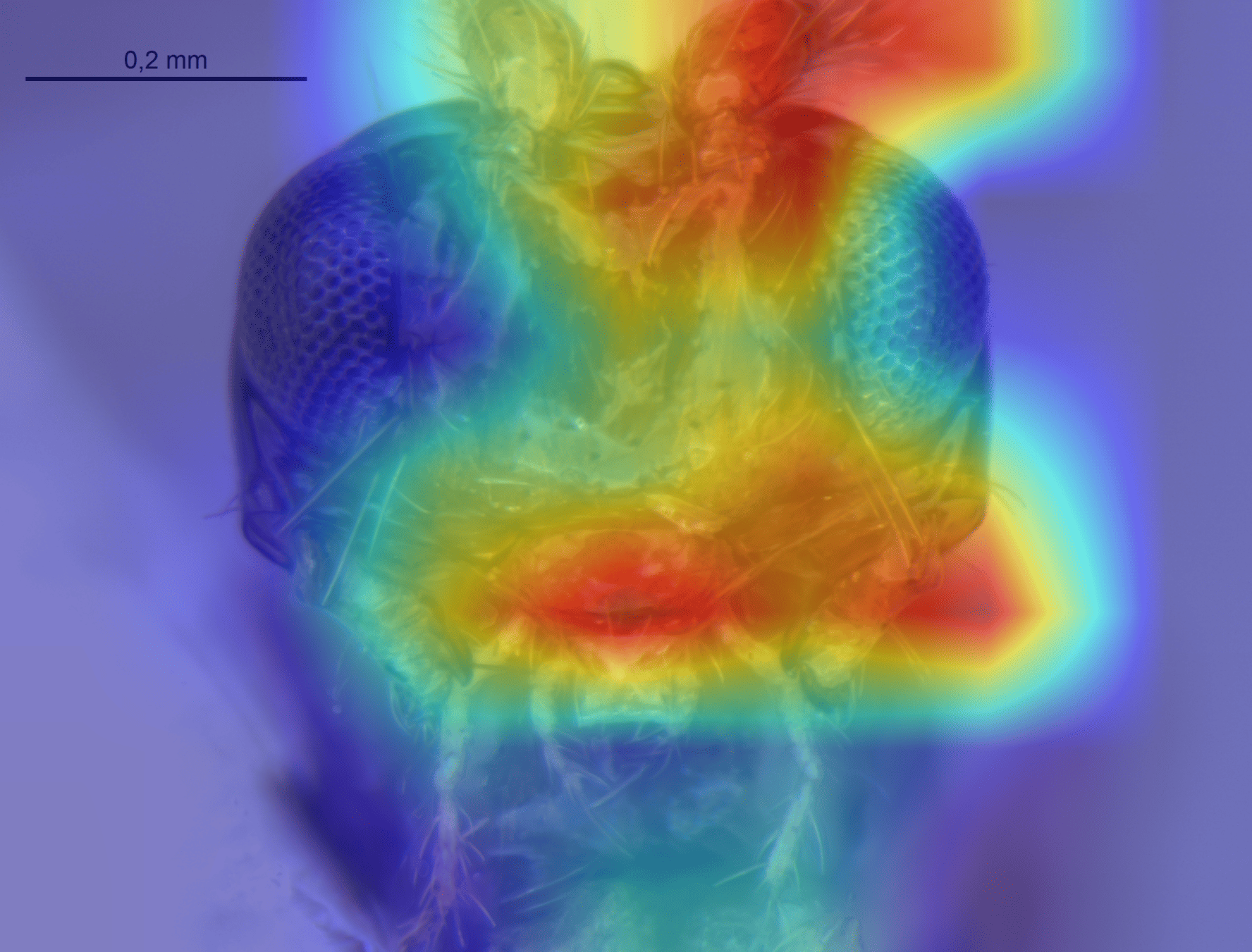}
	\caption{HiResCAM visualizations for Braconidae reveal that the YOLOv26 architecture identifies and prioritizes the nandibles open aligning with established entomological keys.}
	\label{braco_correct_front}
\end{figure}

Furthermore, an essential diagnostic trait is the fusion of metasomal tergites 2 and 3, creating a rigid structural unit that is clearly visible in the lateral profiles of the specimens (Figure \ref{braco_correct_lateral}). At the subfamily level, specialized mandibular structures serve as additional key identifiers. In specific groups, the mandibles are characteristically open and non-overlapping 
~\cite{Sharkey_key_2023,QuickeAsia2023}, as documented in the frontal view provided in Figure \ref{braco_correct_front}

\subsection{Apidae}
\begin{figure}[H]
    \centering
        \includegraphics[width=0.8\linewidth]{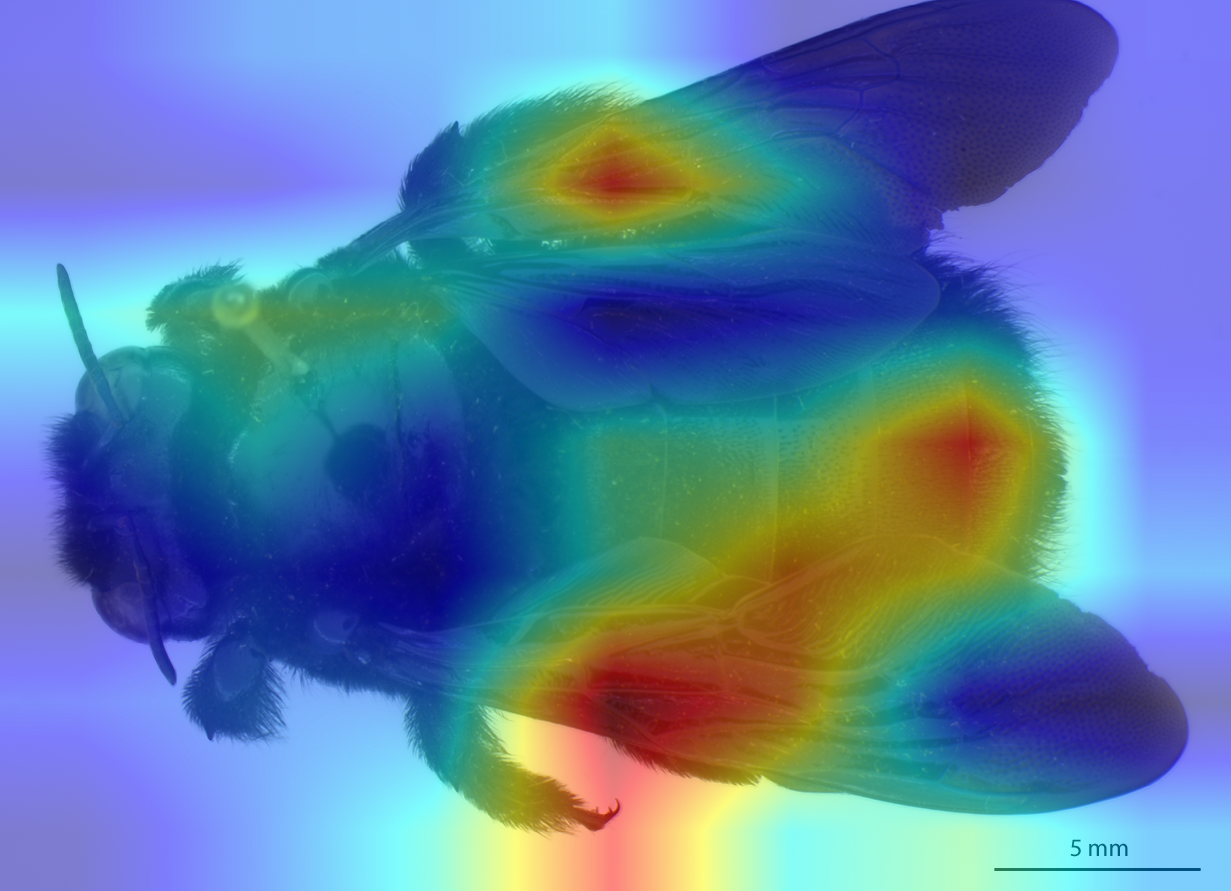}
	\caption{HiResCAM visualizations for Apidae reveal that the YOLOv26 architecture identifies and prioritizes the fully developed complex venation.}
	\label{Apidae_158.png}
\end{figure}

\begin{figure}[H]
    \centering
        \includegraphics[width=0.85\linewidth]{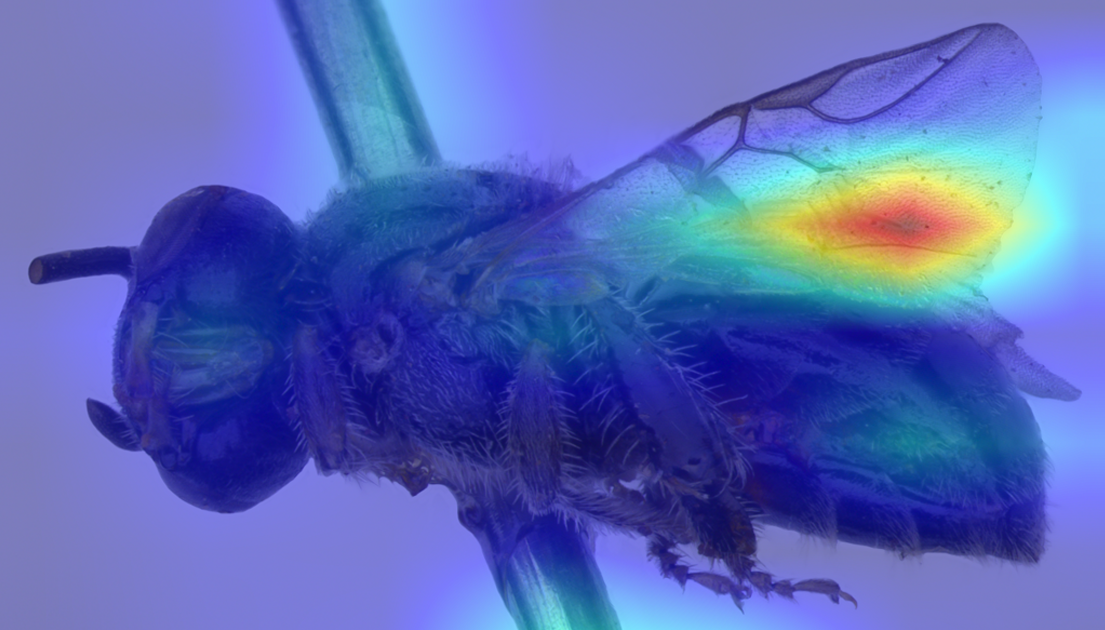}
	\caption{HiResCAM visualizations for Apidae reveal that the YOLOv26 architecture identifies and prioritizes the fully developed complex venation.}
	\label{Apidae_97.png}
\end{figure}

For Apidae identification, the model's performance was evaluated in recognizing body regions and structures that are traditional diagnostic features crucial for the taxonomic classification of various bee tribes. A key example is the variation in wing venation; the model accurately differentiated between the fully developed, complex venation typical of most apid groups (e.g., Figure \ref{Apidae_158.png}) and the significantly reduced or simplified patterns characteristic of stingless bees (e.g., Figure \ref{Apidae_97.png}) ~\cite{michener2007bees}. 

\begin{figure}[H]
    \centering
        \includegraphics[width=0.7\linewidth]{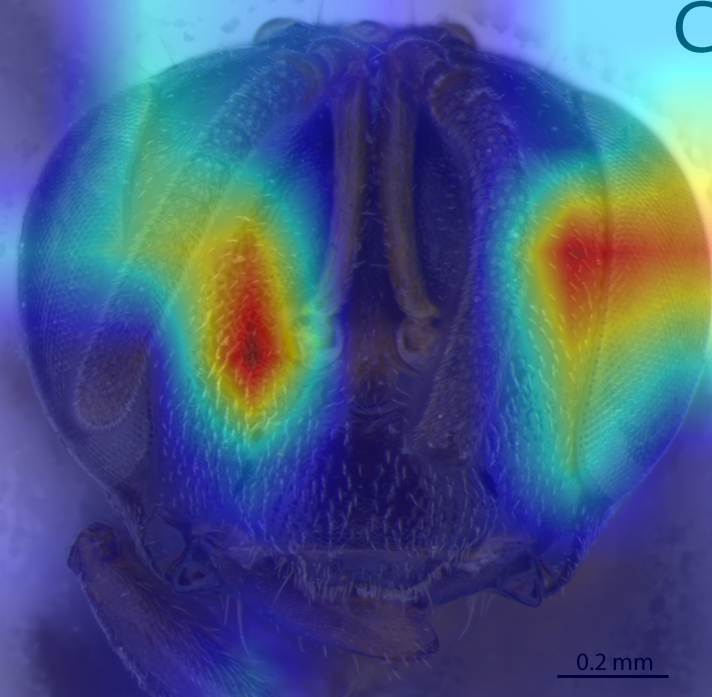}
	\caption{HiResCAM visualizations for Apidae reveal that the YOLOv26 architecture identifies and prioritizes the medial margin of the compound eyes.}
	\label{Apidae_135.png}
\end{figure}

\begin{figure}[ht]
    \centering
        \includegraphics[width=0.8\linewidth]{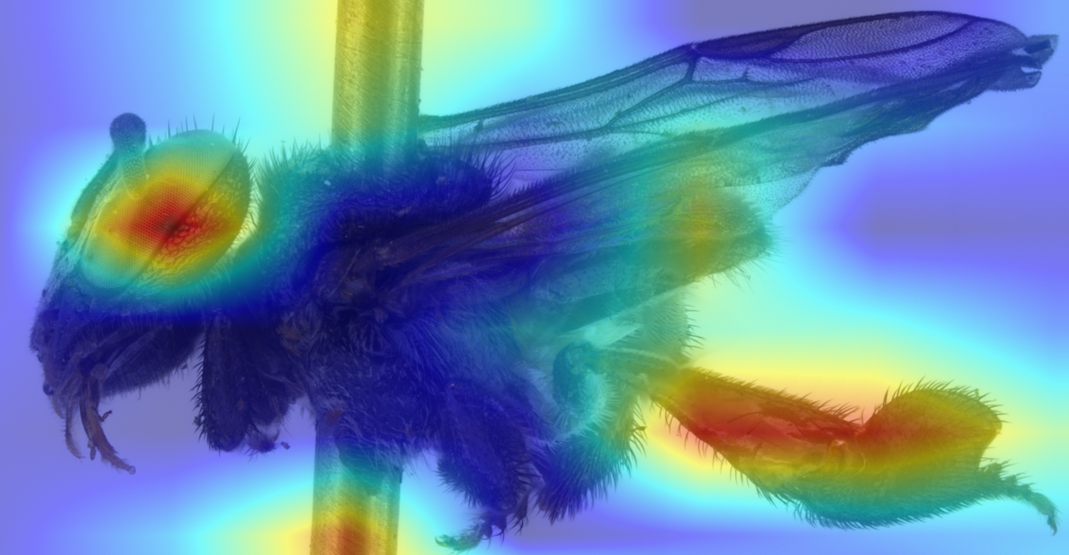}
	\caption{HiResCAM visualizations for Apidae reveal that the YOLOv26 architecture identifies and prioritizes the presence of either a scopa or a corbicula.}
	\label{Apidae_103.png}
\end{figure}

\begin{figure}[ht]
    \centering
        \includegraphics[width=0.8\linewidth]{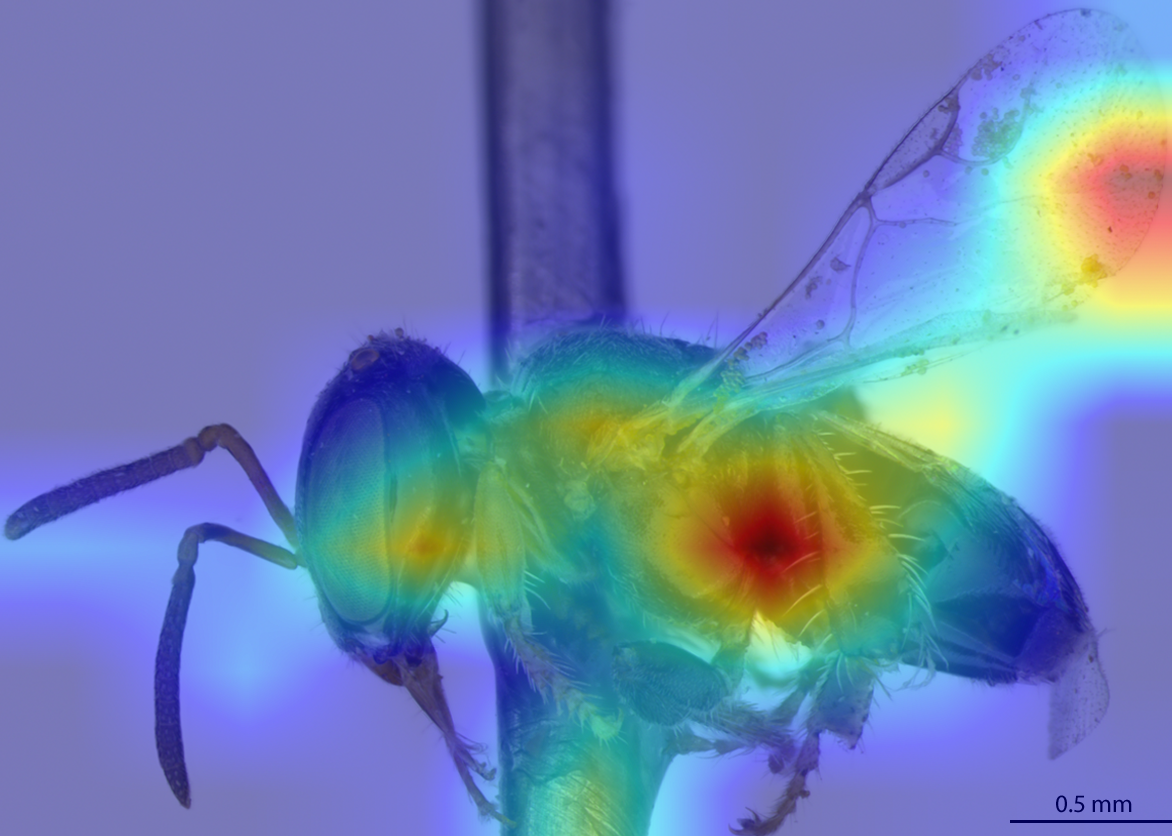}
	\caption{HiResCAM visualizations for Apidae reveal that the YOLOv26 architecture identifies and prioritizes presence of either a scopa or a corbicula.}
	\label{Apidae_143.png}
\end{figure}

Additionally, the model captured important head features, particularly the medial margin of the compound eyes and the relative proportions of the different regions (e.g., Fig \ref{Apidae_135.png}), which are diagnostic at the genus level. Another key area highlighted by the model was the hind leg, with a focus on specialized structures involved in pollen transport (Figure \ref{Apidae_143.png}). The presence of either a scopa or a corbicula (Figure \ref{Apidae_103.png}) is a decisive trait for distinguishing tribes of Apidae ~\cite{michener2007bees}.

Although the primary focus of this proposed framework is the taxonomic triage of the Ichneumonoidea superfamily, extending the visual interpretability analysis to out-group families such as Apidae is essential to demonstrate the model's broad morphological generalization capabilities. 
% Evaluating a distinct group with specialized functional morphology allows us to verify that the network effectively maps true taxonomic boundaries rather than merely memorizing dataset-specific artifacts.

% \begin{figure}[ht]
%     \centering
%         \includegraphics[width=0.7\linewidth]{figs/Apidae_136.png}
% 	\caption{HiResCAM visualizations for Apidae reveal that the YOLOv26 architecture identifies and prioritizes the medial margin of the compound eyes.}
% 	\label{Apidae_136.png}
% \end{figure}

\section{Conclusion}
\label{sec:conclusion}
This study presented a YOLO-based deep learning framework for automated identification of Ichneumonoidea wasps. The proposed system achieved strong classification performance while maintaining computational efficiency suitable for real-time or near real-time applications.

The incorporation of Grad-CAM provided critical insight into the model's decision-making process. Visualization results indicate that the network consistently attends to biologically meaningful morphological features, including wing venation patterns, antennal morphology, and metasomal segmentation. This alignment between learned representations and taxonomic traits enhances model credibility and supports its applicability in scientific workflows.

From a practical standpoint, rather than replacing human expertise, the proposed framework serves as an automated triage system that efficiently routes bulk samples to the appropriate taxonomic specialists based on family-level predictions. By eliminating the manual burden of preliminary sorting, it optimizes human resources and allows expert taxonomists to concentrate their efforts on more complex, fine-grained classifications, such as genus- or species-level identification. Consequently, this approach offers scalable support for biodiversity surveys, ecological research, and biological control initiatives. Moreover, the explainability component addresses a major limitation of black-box deep learning systems by enabling the qualitative validation of predictions through biologically relevant morphological features.

Future work may extend this framework to subfamily or genus identification and incorporate larger, more diverse datasets to improve generalization. Integration into mobile or field-deployable systems would further increase accessibility and real-world utility.

% Overall, the study demonstrates that combining modern object detection models with explainable AI techniques provides a robust and interpretable solution for automated insect taxonomy.

\bibliographystyle{IEEEtran}
\bibliography{references}

\begin{IEEEbiography}[{\includegraphics[width=1in,height=1.25in,clip,keepaspectratio]{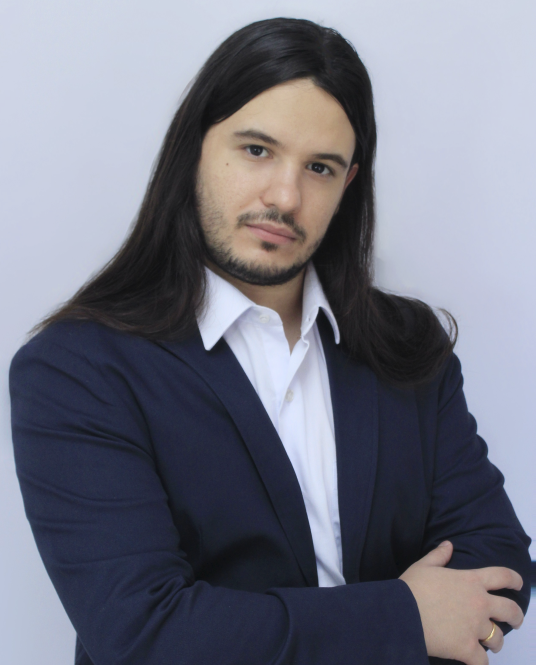}}]{João Manoel Herrera Pinheiro}
received the M.Sc. degree in Mechanical Engineering, with a focus on computer vision and robotics, and the B.E. degree in Mechatronics Engineering from the University of São Paulo (USP). He is currently pursuing the D.Sc. degree in Electrical Engineering, with research interests centered on computer vision, machine learning, and image and signal processing. He has also completed two postgraduate specialization programs: Didactic–Pedagogical Processes for Distance Learning (UNIVESP) and MBA in Software Engineering (USP). He serves as a reviewer for several international journals, including IEEE Latin America Transactions, Scientific Data (Nature), Artificial Intelligence (IBERAMIA), and the Journal of the Brazilian Computer Society.
\end{IEEEbiography}

\begin{IEEEbiography}[{\includegraphics[width=1in,height=1.25in,clip,keepaspectratio]{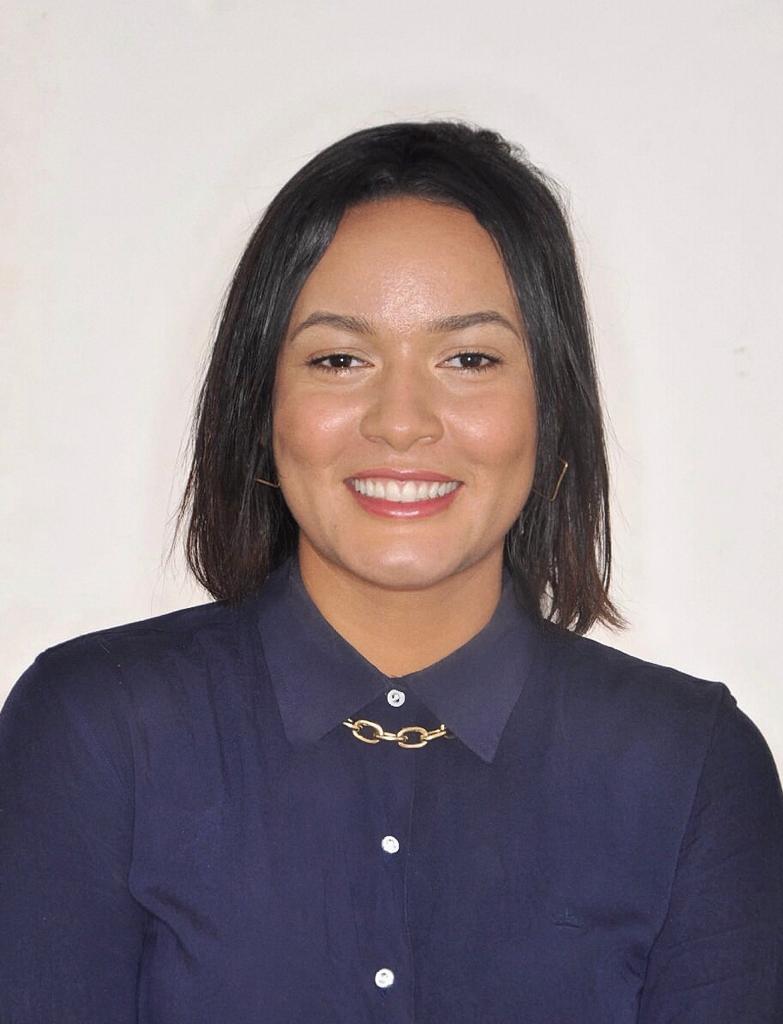}}]{Gabriela do Nascimento Herrera}
received the B.E. degree in Agricultural Engineering and the M.Eng. degree in Agronomy, with a focus on Hymenoptera parasitoids, from the Federal University of Acre (UFAC). She is currently pursuing the D.Sc. degree in Ecology and Natural Resources at the Federal University of São Carlos (UFSCar), with a research focus on the Ichneumonoidea superfamily in the Western Amazon rainforest and biological control.
\end{IEEEbiography}

\begin{IEEEbiography}[{\includegraphics[width=1in,height=1.25in,clip,keepaspectratio]{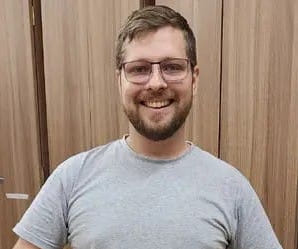}}]{Alvaro Doria dos Santos}
received the B.Sc. degree in Biological Sciences from Mackenzie Presbyterian University in 2015, the M.Sc. and the Ph.D degree in Systematics, Animal Taxonomy and Biodiversity from the Museum of Zoology of the University of São Paulo,  in 2024. He is currently a Postdoctoral Researcher at the Federal University of Tocantins (UFT), Brazil. His research interests include the taxonomy and evolution of Darwin wasps (Hymenoptera: Ichneumonidae), insect systematics, phylogenetics, biodiversity and biological collection management and digitization.
\end{IEEEbiography}

\begin{IEEEbiography}[{\includegraphics[width=1in,height=1.25in,clip,keepaspectratio]{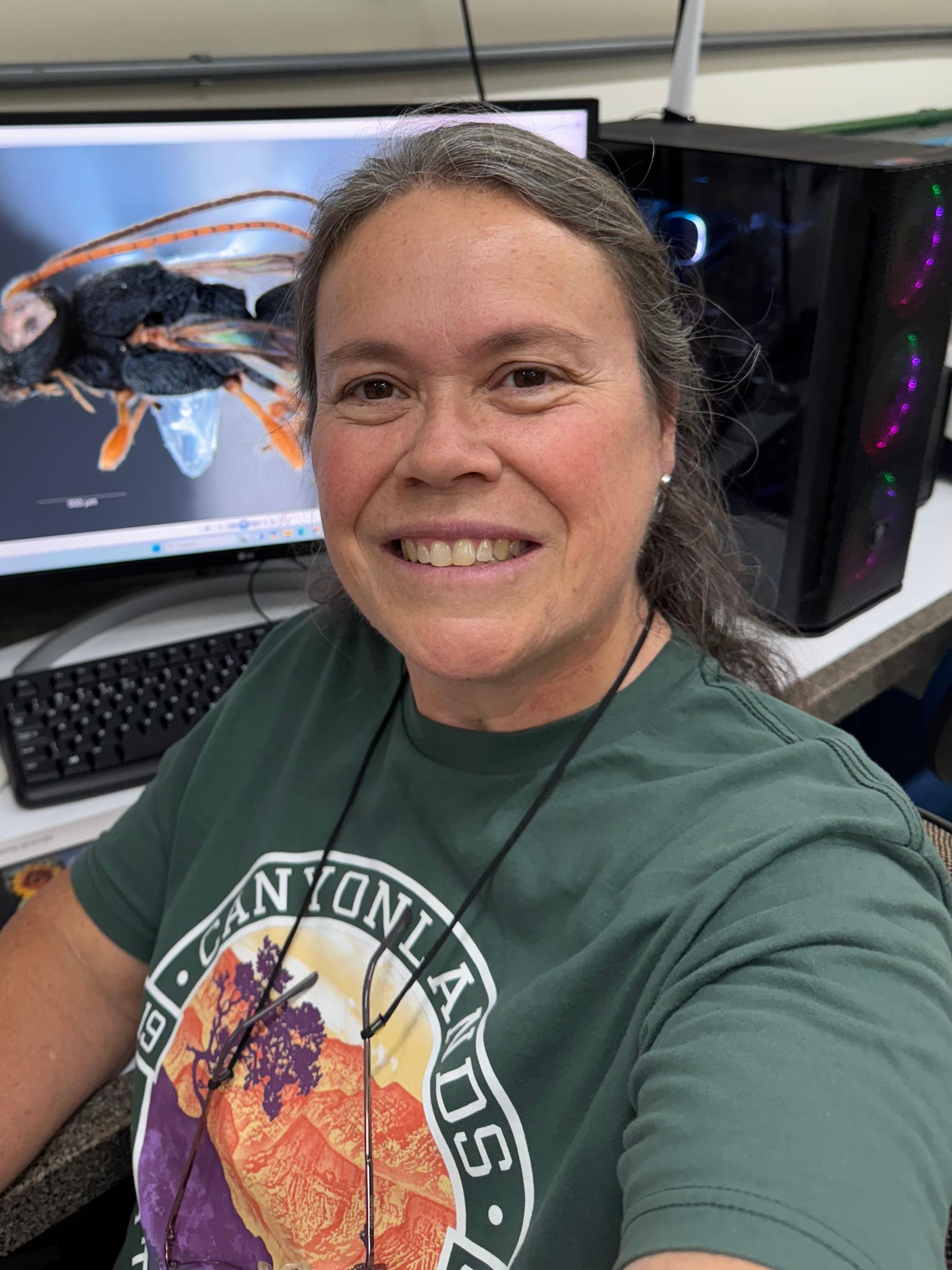}}]{Luciana Bueno dos Reis Fernandes}
received the Bachelor's and Licentiate Degree in Biological Sciences from the Federal University of São Carlos (UFSCar) in 1993, and both her M.Sc. (1999) and Ph.D. (2003) in Sciences from the Graduate Program in Ecology and Natural Resources at UFSCar, with research focused on the biology and taxonomy of parasitoid Hymenoptera and the bionomics of Lepidoptera. She currently works as a Biologist in the Department of Ecology and Evolutionary Biology at UFSCar, where she is responsible for the acquisition and processing of photographs for scientific research and publications.
\end{IEEEbiography}

\begin{IEEEbiography}
[{\includegraphics[width=1in,height=1.25in,clip,keepaspectratio]{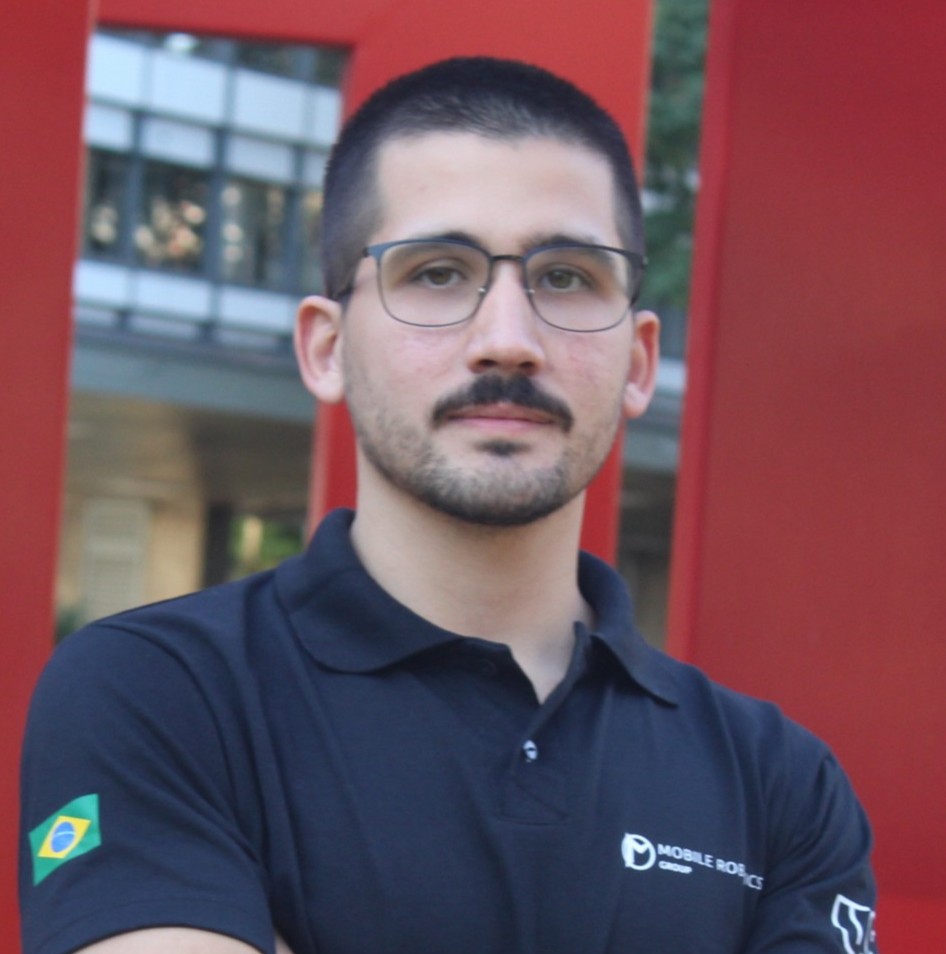}}]{Ricardo V. Godoy} received the Bachelor of Engineering in Mechatronics Engineering in 2019, followed by M.Sc. in 2021 in Mechanical Engineering from the University of São Paulo, São Carlos, Brazil. He received his PhD in Mechatronics Engineering with the New Dexterity Research Group of the University of Auckland, New Zealand, where he worked on the analysis and development of novel Human-Machine Interfaces (HMI) for the control of robotic and bionic devices while focusing on the challenges and limitations in the use of HMI for robust grasping and decoding of dexterous, in-hand manipulation tasks. He is currently pursuing his postdoc at the University of São Paulo, São Carlos, Brazil, working towards the development of robotic frameworks for inspection and automation, focusing on manipulation and loco-manipulation frameworks.
\end{IEEEbiography}

\begin{IEEEbiography}[{\includegraphics[width=1in,height=1.25in,clip,keepaspectratio]{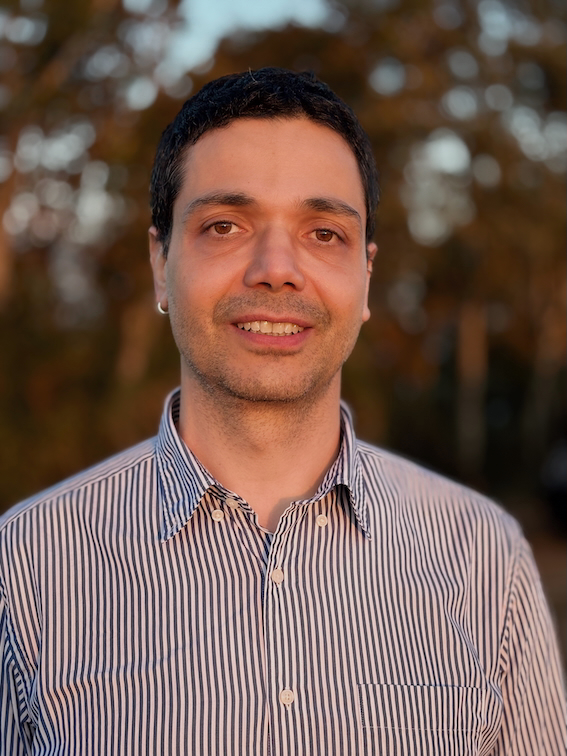}}]{Eduardo A. B. Almeida}
received the B.Sc. degree in Biology and the M.Sc. in Ecology from the Universidade Federal de Minas Gerais (Brazil) in 2000 and 2002, respectively, and his PhD degree in Entomology from Cornell University (USA) in 2007. He is currently an Associate Professor at the Universidade de São Paulo (Brazil), where he coordinates a research group in comparative biology of insects, with emphasis on bees. His research interests include taxonomy, phylogenetics, biogeography, and comparative morphology.
\end{IEEEbiography}

\begin{IEEEbiography}[{\includegraphics[width=1in,height=1.25in,clip,keepaspectratio]{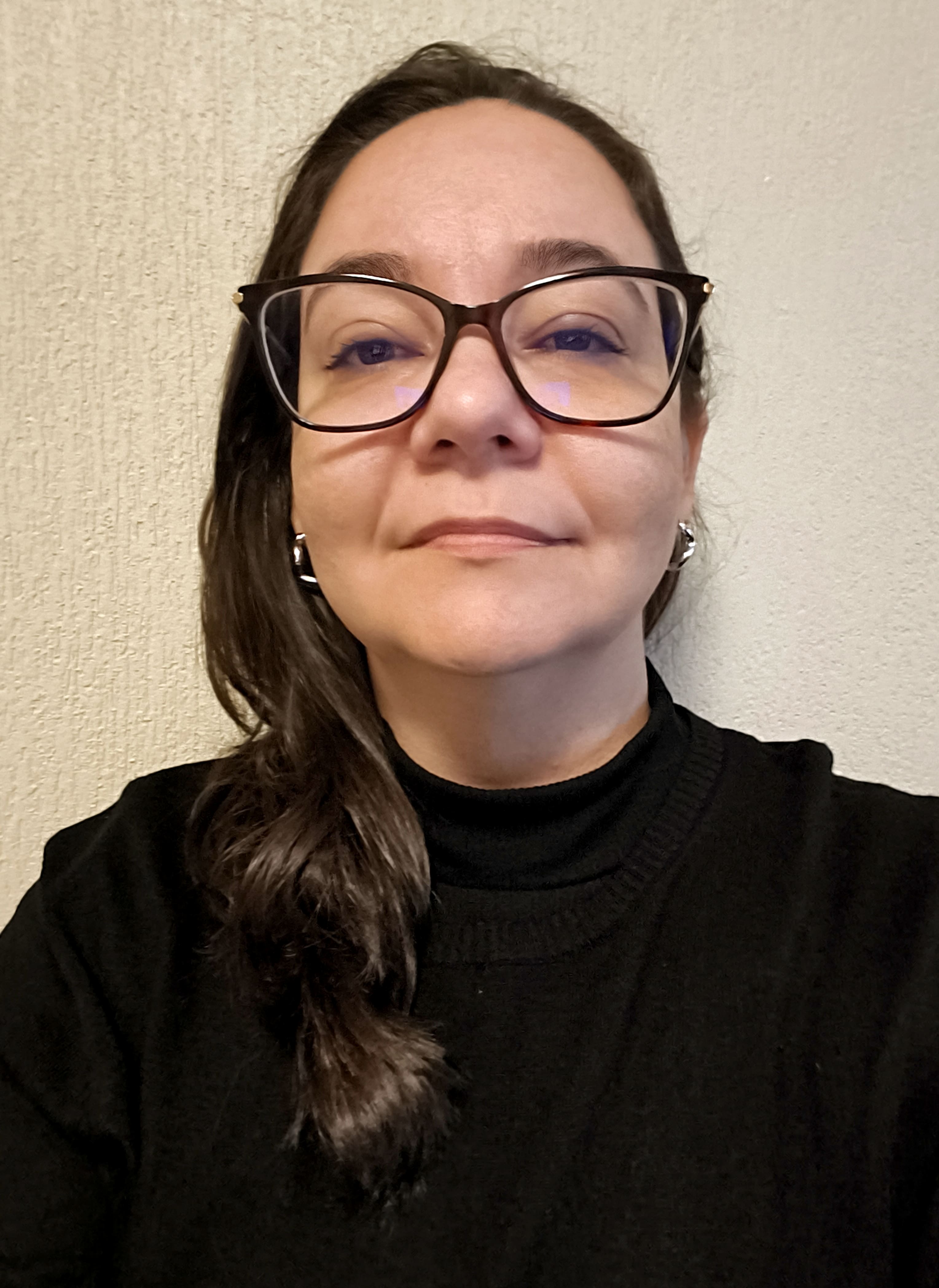}}]{Helena Carolina Onody}
received the B.Sc. and Licentiate degrees in Biological Sciences and the Ph.D. degree in Ecology and Natural Resources from the Federal University of São Carlos (UFSCar), Brazil. She is currently a Professor at the State University of Piauí (UESPI), Brazil and member of the Program in Biodiversity and Conservation at the Federal University of Piauí (UFPI). Her research interests include insect taxonomy and systematics, particularly parasitoid wasps (Hymenoptera: Ichneumonidae), biodiversity, entomological collections, biological collection management and digitization, and insect ecology.
\end{IEEEbiography}

\begin{IEEEbiography}[{\includegraphics[width=1in,height=1.25in,clip,keepaspectratio]{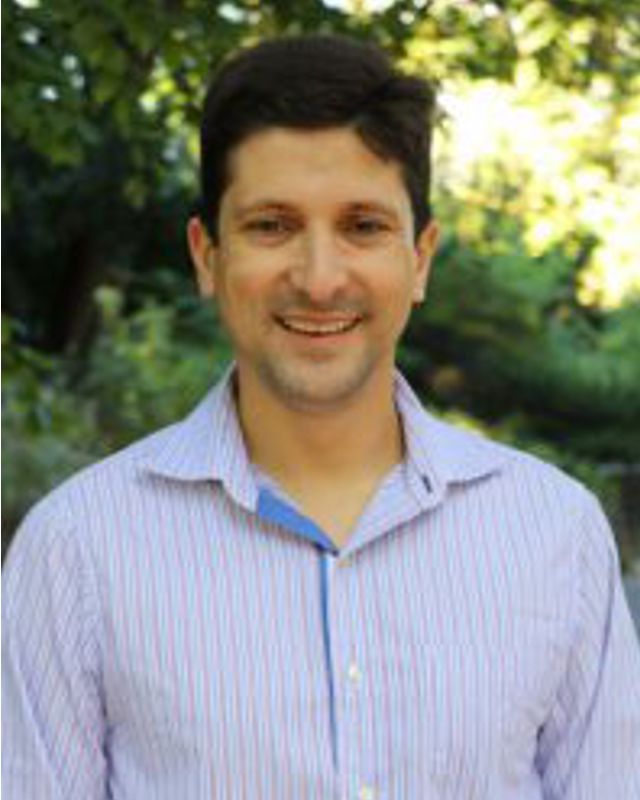}}]{Marcelo Andrade da Costa Vieira}
received the B.Sc. degree in Electrical Engineering from the University of São Paulo, Brazil, in 1996. He received the M.Sc. and Ph.D. degrees in Electrical Engineering from the same institution in 1999 and 2005, respectively. He was a Postdoctoral Research Fellow with the University of Pennsylvania, USA, from 2012 to 2013. He is currently an Associate Professor at the University of São Paulo, where he coordinates the Laboratory of Advanced Vision and Imaging (LAVI) and leads the Computer Vision and Medical Image Processing Research Group. His research interests include digital mammography, digital breast tomosynthesis, magnetic resonance imaging, medical image processing, noise modeling, image denoising, artificial intelligence, radiation dose reduction, and task-based assessment of image quality.
\end{IEEEbiography}

\begin{IEEEbiography}[{\includegraphics[width=1in,height=1.25in,clip,keepaspectratio]{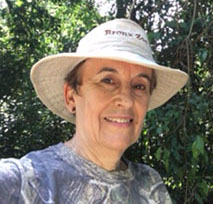}}]{Angélica Maria Penteado-Dias}
was undergraduates in Natural History from the School of Philosophy, Sciences and Letters of Rio Claro from São Paulo State University (UNESP) Brazil, in 1972, the M.Sc. degree in Zoology in 1976, and the Ph.D. degree in Zoology in 1981, both from the University of São Paulo (USP), Brazil. She is currently a Full Professor with the Federal University of São Carlos (UFSCar), where she also coordinates several research projects. Her research interests include the taxonomy, systematics, diversity, and evolution of parasitoid wasps, with emphasis on Hymenoptera, particularly Braconidae and Ichneumonidae.
\end{IEEEbiography}

\begin{IEEEbiography}[{\includegraphics[width=1in,height=1.25in,clip,keepaspectratio]{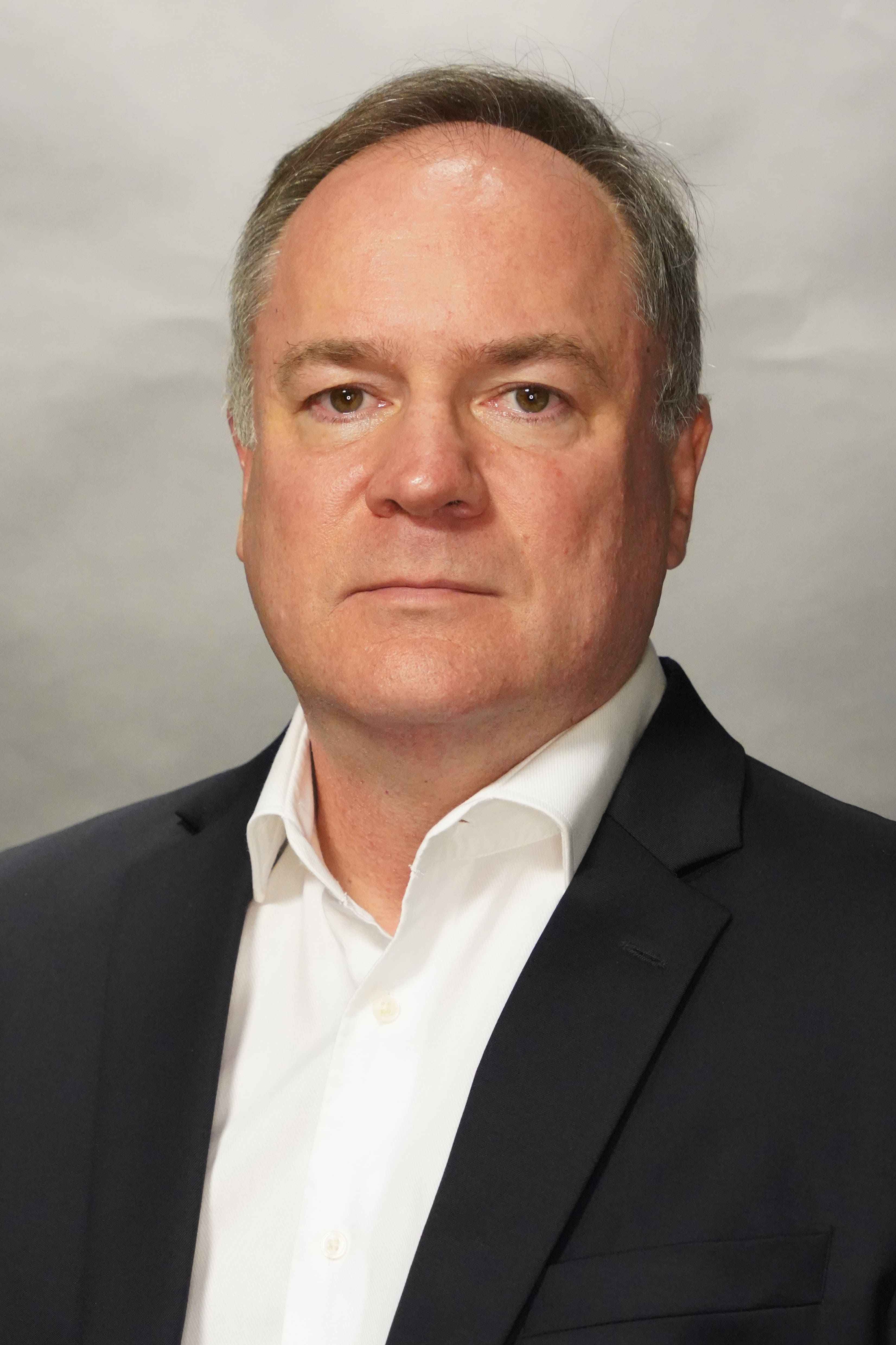}}]{Marcelo Becker} received the B.Sc. degree in Mechanical Engineering (Mechatronics) from the University of São Paulo, Brazil, in 1993, and the M.Sc. and D.Sc. degrees in Mechanical Engineering from the University of Campinas, Brazil, in 1997 and 2000, respectively. He was a visiting researcher at ETH Zürich and did a sabbatical at EPF Lausanne, Switzerland. He is currently an Associate Professor at the University of São Paulo and coordinates the Mobile Robotics Group and the USP Center of Robotics (CRob). His research interests include mobile robotics, automation, perception systems, and mechatronic design for applications in agriculture and industrial automation.
\end{IEEEbiography}

%If you do not have or do not want to include a photo, you can use IEEEbiographynophoto as shown below:
\EOD

\end{document}